\documentclass[twoside]{article}

\usepackage{arxiv}

\usepackage{bm}
\usepackage{graphicx}
\usepackage{amsmath,amssymb,mathtools,amsfonts,amsthm}
\usepackage{booktabs}
\usepackage{multirow}
\usepackage{longtable}
\usepackage{array}
\usepackage[shortlabels]{enumitem}
\usepackage[margin=2.5cm]{geometry}
\usepackage[english]{babel}
\usepackage[normalem]{ulem}
\usepackage{soul}
\usepackage{float}
\usepackage{textcomp}
\usepackage{verbatim}
\usepackage{setspace}
\usepackage{subcaption}
\usepackage{mwe}       
\usepackage{lineno}
\usepackage{forest}
\usepackage{algorithm}
\usepackage{algorithmic}
\usepackage{tikz}
\usepackage{pgfplots}
\usepackage{tikz-cd}
\usepackage[percent]{overpic}
\usetikzlibrary{bayesnet, arrows.meta}
\usetikzlibrary{backgrounds}
\pgfdeclarelayer{background}
\pgfsetlayers{background,main}
\newtheorem{proposition}{Proposition}

\usepackage[hidelinks]{hyperref}

\usepackage[round]{natbib}

\pgfplotsset{compat=1.18}

\begin{document}

\twocolumn[

\aistatstitle{Meta-probabilistic Modeling}

\aistatsauthor{ \large Kevin Zhang$^1$ \And \large Yixin Wang$^2$ }

 \aistatsaddress{ } 

]

\begin{abstract}
    Probabilistic graphical models (PGMs) are widely used to discover latent structure in data, but their success hinges on selecting an appropriate model design. In practice, model specification is difficult and often requires iterative trial-and-error. This challenge arises because classical PGMs typically operate on individual datasets. In this work, we consider settings involving collections of related datasets and propose \emph{meta-probabilistic modeling (MPM)} to learn the generative model structure itself. MPM uses a hierarchical formulation in which global components encode shared patterns across datasets, while local parameters capture dataset-specific latent structure. For scalable learning and inference, we derive a tractable VAE-inspired surrogate objective together with a bi-level optimization algorithm. Our methodology supports a broad class of expressive probabilistic models and has connections to existing architectures, such as Slot Attention. Experiments on object-centric representation learning and sequential text modeling demonstrate that MPM effectively adapts generative models to data while recovering meaningful latent representations.
\end{abstract}

\section{INTRODUCTION} \label{sec:intro}

Probabilistic graphical models (PGMs) provide a principled method for discovering and analyzing latent structure in data~\citep{Silva2009TheHL,Li2013SurveyOP}. PGMs represent the underlying generative process using graphical structures that encode dependencies among random variables, which can enable efficient learning and inference~\citep{Jordan1999AnIT, Wainwright2008GraphicalME}. This framework encompasses a broad family of models, ranging from latent variable models for hierarchical organization~\citep{Bishop1998HierarhicalLVM} to state-space models for capturing temporal dynamics~\citep{rabiner1986HMM, pmlr-v80-doerr18a}. Because of their expressiveness and interpretability, PGMs are widely applied across numerous domains, including semantic topic discovery in natural language processing~\citep{Blei2003LatentDA,Blei2006DynamicTM,Blei2012ProbabilisticTM} and molecular interaction modeling in computational biology~\citep{Airoldi2007GettingSI}.

\begin{figure}[b!]
    \centering
    \includegraphics[width=\linewidth]{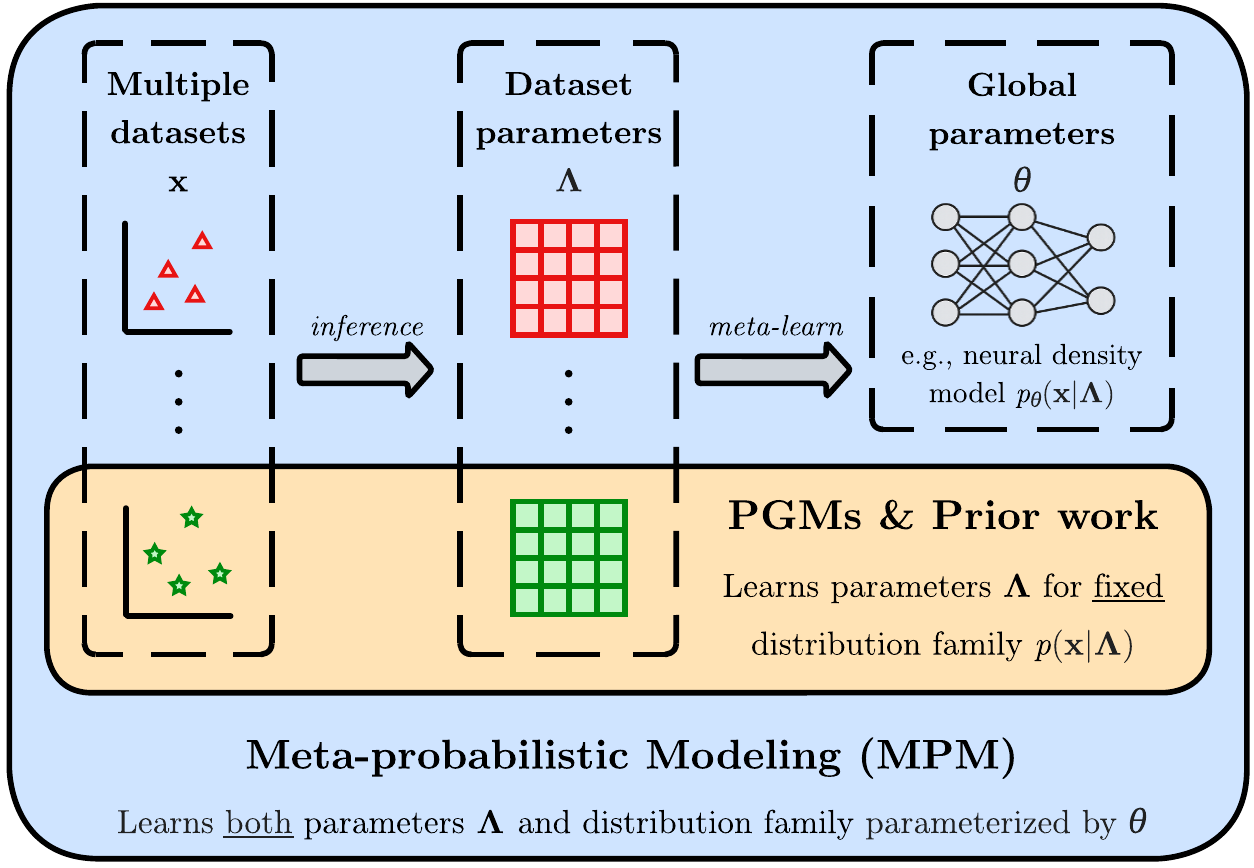}
    \caption{MPM infers dataset parameters $\Lambda$ from multiple datasets $\mathbf{x}$, and meta-learns the conditional distribution $p_{\theta}(\mathbf{x} \mid \Lambda)$ using a (potentially neural) parameterization. Unlike classical PGMs and most prior work, which typically assume a fixed model family (e.g., Gaussian), MPM jointly learns both the distribution family and its parameters from data.}
    \label{fig:mpm_overview}
\end{figure}

However, the effectiveness of PGMs depends on selecting well-specified models that appropriately capture the underlying dependencies and structure of the data~\citep{Koller2009ProbabilisticGM}. This typically requires the practitioner to iteratively refine both the graphical structure and the associated distributional families. Designing the model by hand rather than learning it from data can lead to misspecification, including imprecise modeling of observations and latent components, rigid graphical topologies that fail to adapt to heterogeneous data, and structural assumptions that may be overly restrictive~\citep{Juang1985MixtureARHMM,Kingma2013AutoEncodingVB}.

\textbf{Meta-probabilistic modeling.} To address these challenges, we introduce \emph{meta-probabilistic modeling (MPM)}, an approach for learning the model directly from data. Because PGMs usually operate at the level of a single dataset, the underlying form is under-specified. Our approach instead leverages a collection of datasets sharing related latent structure to infer the model, as shown in Figure~\ref{fig:mpm_overview}.

We posit a hierarchical architecture with \emph{global parameters} shared across datasets and \emph{dataset-specific parameters} that capture variation at the dataset and local level. The global components specify the form of the distributional families and are parameterized flexibly (e.g., using neural networks), while the local components model the underlying latent structure. This design combines the interpretability of probabilistic modeling with the expressive capacity of deep learning architectures. 

As with most latent variable models, computing the posterior distribution over the model parameters in our method is intractable and must be approximated~\citep{Wainwright2008GraphicalME,Koller2009ProbabilisticGM}. We show that inference in meta-probabilistic modeling can remain tractable, even when parts of the generative process are parameterized by a complex family, such as neural networks. At a high level, we construct a surrogate potential inspired by recognition networks in variational autoencoders (VAEs), which enables analytic local coordinate ascent updates for dataset-specific parameters, while learning the global generative components via gradient-based optimization.

\textbf{Contributions.} Our main contributions are as follows: (1) we propose \emph{meta-probabilistic modeling (MPM)}, a probabilistic framework that learns generative model structure across multiple related datasets by combining hierarchical PGMs with flexible global  parameterizations; (2) we derive an efficient and scalable learning algorithm with principled connections to Slot Attention~\citep{Locatello2020ObjectCentricLearning}; and (3) we validate MPM on object-centric representation learning and sequential text modeling, where it recovers meaningful latent structure while flexibly adapting to complex data.

Note that while we parameterize the shared components using neural networks, our method itself is agnostic to the specific architectures employed. Rather, our primary aim in this work is to articulate a broad class of rich, probabilistic models for multi-dataset settings and to derive a concrete learning and inference procedure.

\section{META-PROBABILISTIC MODELING} \label{sec:mpm}

In this section, we formalize MPM and our corresponding learning and inference algorithm.


\subsection{Problem formulation}

\begin{figure}[t]
    \centering
    \begin{tikzpicture}
    
    \node[latent] (z) {$z$};
    \node[obs, below=of z] (x) {$x$};
    \node[latent, left=of x] (lambda) {$\lambda$};
    \node[latent, right=of x] (theta) {$\theta$};
    \node[latent, left=of lambda, xshift=0.3 cm] (eta) {$\eta$};
    
    \edge {lambda} {z};
    \edge {lambda} {x};
    \edge {z} {x};
    \edge {theta} {x};
    \edge {eta} {lambda};
    
    \plate {Nplate} {(z)(x)} {$N$};
    \plate[inner sep=0.5cm, xshift=0.2cm, yshift=-0.2cm] {Mplate} {(lambda)(z)(x)} {};
    \node[anchor=south east, inner sep=5pt] at (Mplate.south east) {$M$};
    
    \end{tikzpicture}
    \caption{The latent variable meta-probabilistic model separates global and dataset parameters hierarchically: $\theta$ indexes the data generating distribution family $p_\theta(x \mid z, \lambda)$, and $\eta$ defines a prior over $\lambda$.
    }
    \label{fig:mpm_graph}
\end{figure}

\begin{figure*}[t!]
    \centering
    \newcommand{\myrightarrow}{\xrightarrow{\phantom{........}}}

    \begin{subfigure}[t]{0.32\textwidth}
        \centering
        \begin{tikzpicture}
        \node (left) at (0,0) {
            \includegraphics[width=0.46\textwidth]{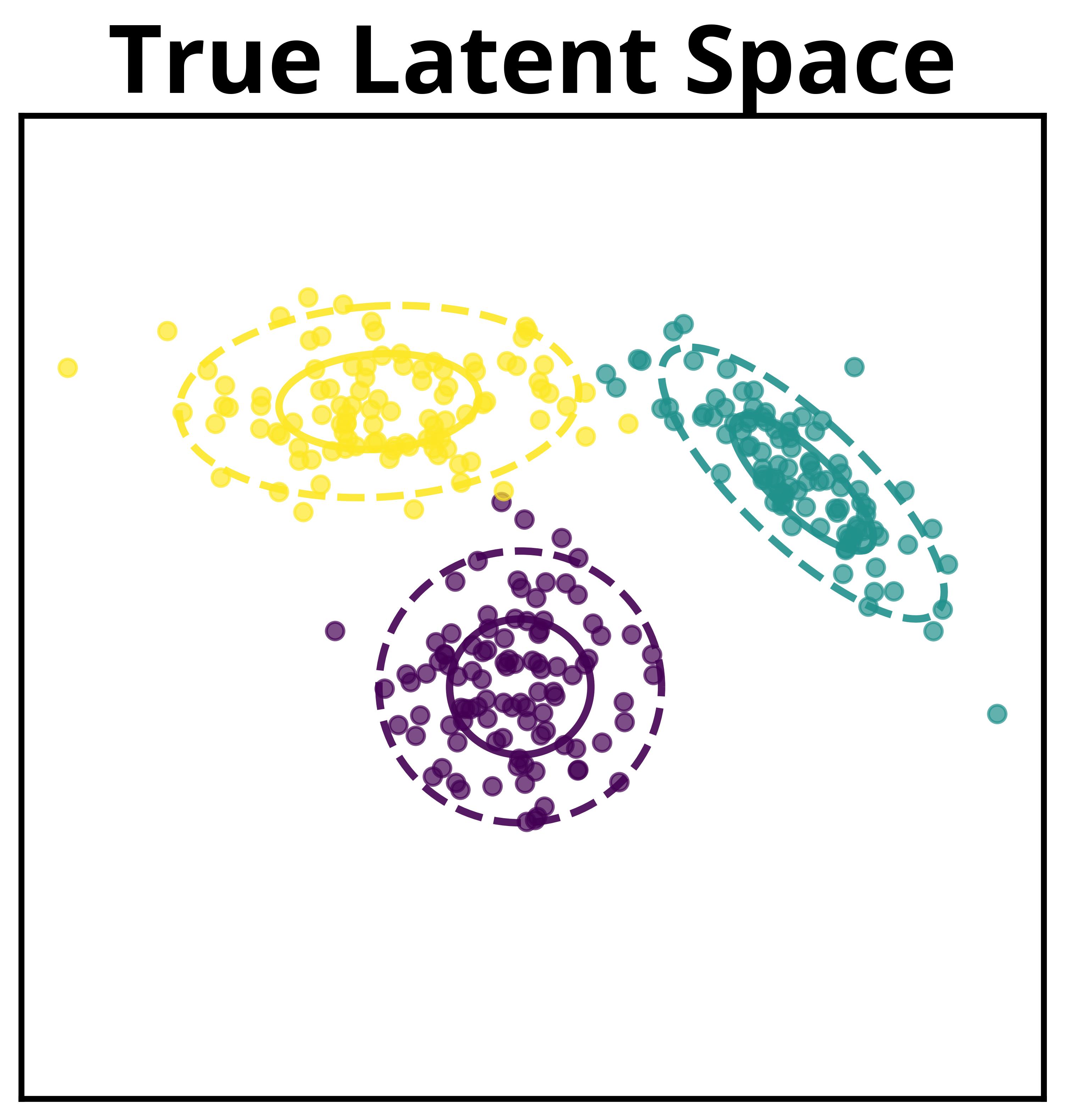}
        };
        \node (right) at (0.49\textwidth,0) {
            \includegraphics[width=0.46\textwidth]{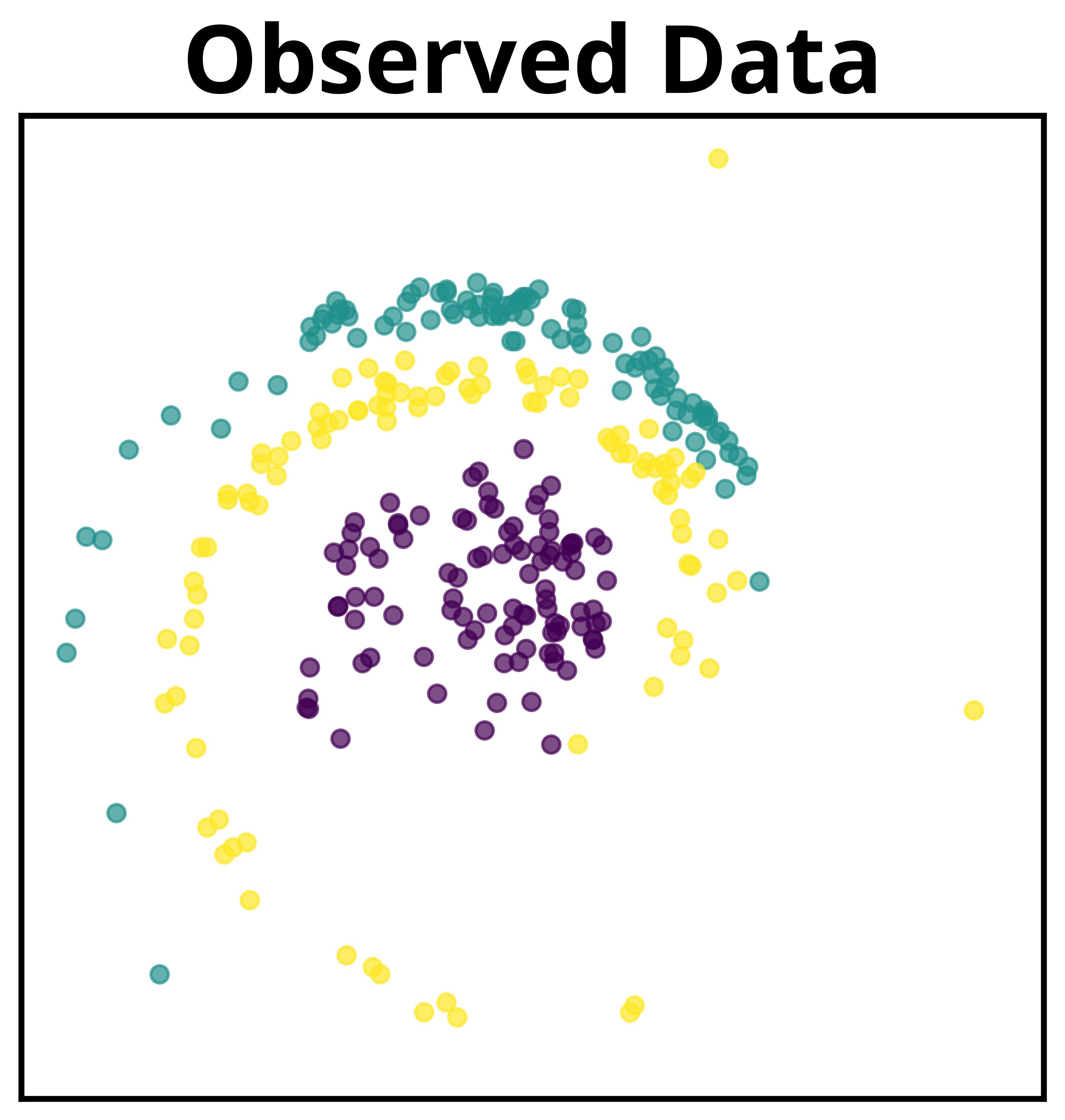}
        };
        \node[anchor=south]
        at ($(left.south east)!0.5!(right.south west) + (-0.18cm,0.15cm)$)
        {$T \myrightarrow$};
        \end{tikzpicture}
        \caption{}
        \label{fig:toy_true}
    \end{subfigure}
    \hfill 
    \begin{subfigure}[t]{0.32\textwidth}
        \centering
        \begin{tikzpicture}
        \node (left) at (0,0) {
            \includegraphics[width=0.46\textwidth]{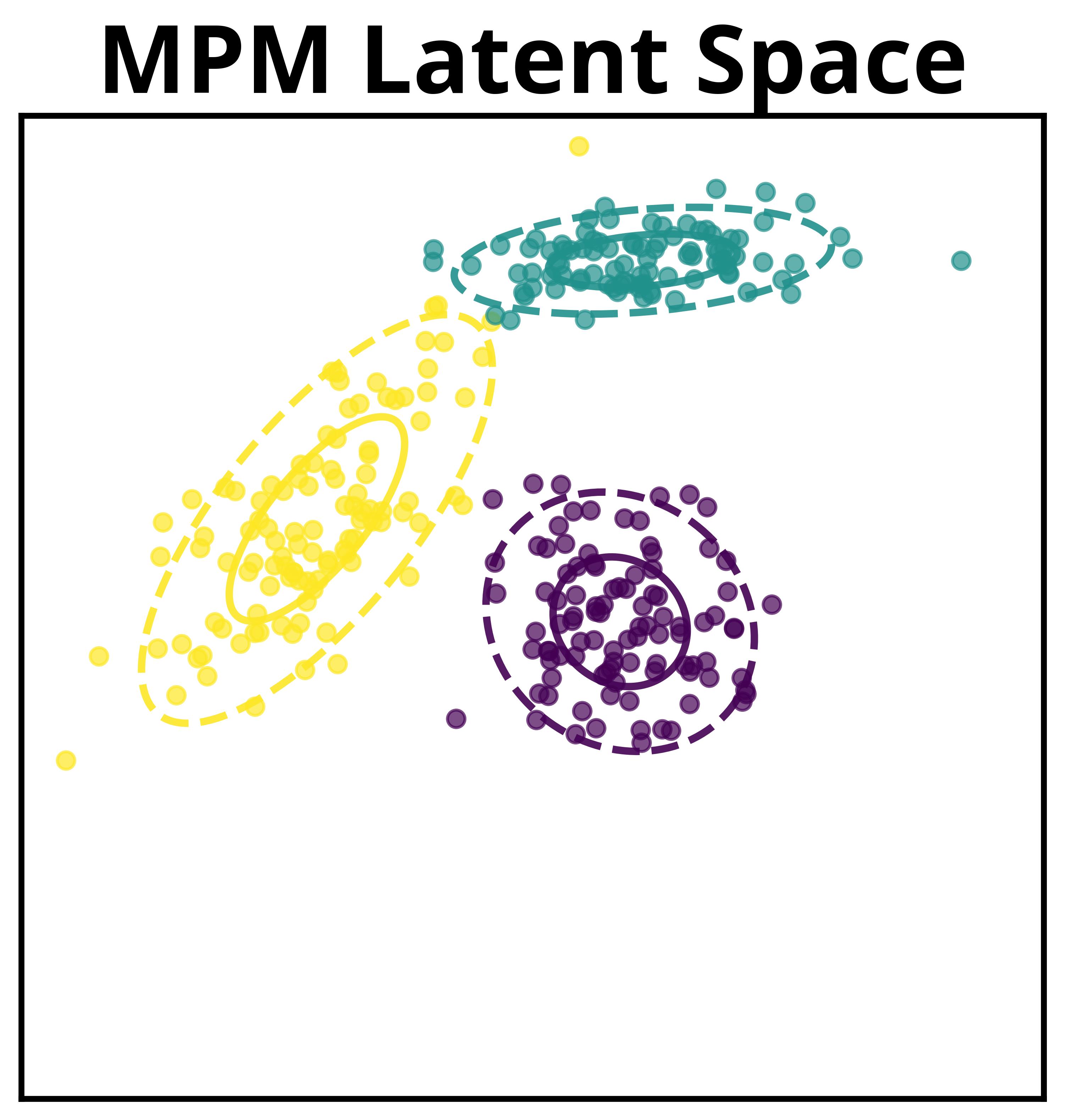}
        };
        \node (right) at (0.49\textwidth,0) {
            \includegraphics[width=0.46\textwidth]{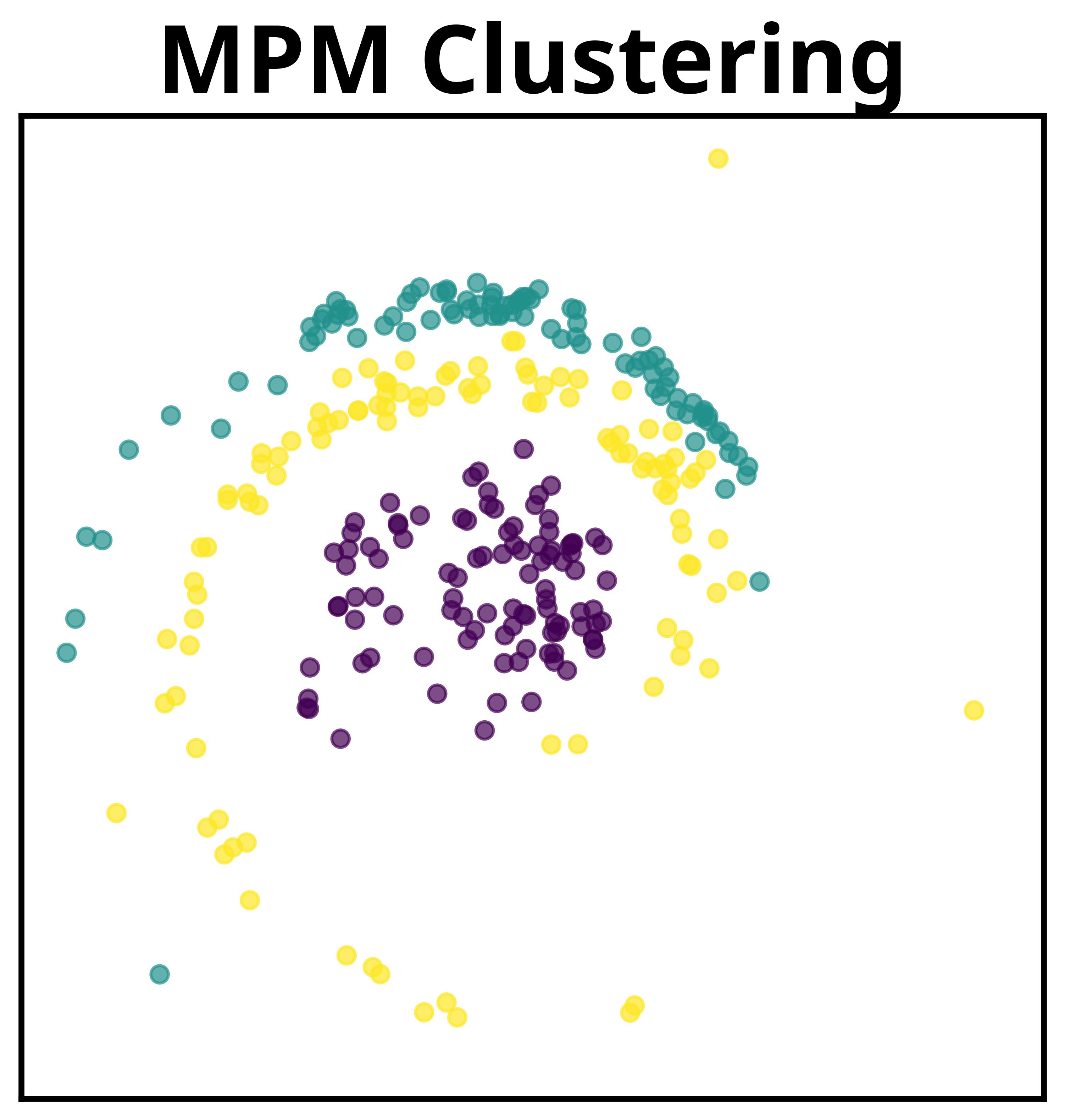}
        };
        \node[anchor=south]
        at ($(left.south east)!0.5!(right.south west) + (-0.18cm,0.15cm)$)
        {$\widehat{T} \myrightarrow$};
        \end{tikzpicture}
        \caption{}
        \label{fig:toy_mpm}
    \end{subfigure}
    \hfill
    \begin{subfigure}[t]{0.32\textwidth}
        \centering
        \begin{tikzpicture}
        \node (left) at (0,0) {
            \includegraphics[width=0.46\textwidth]{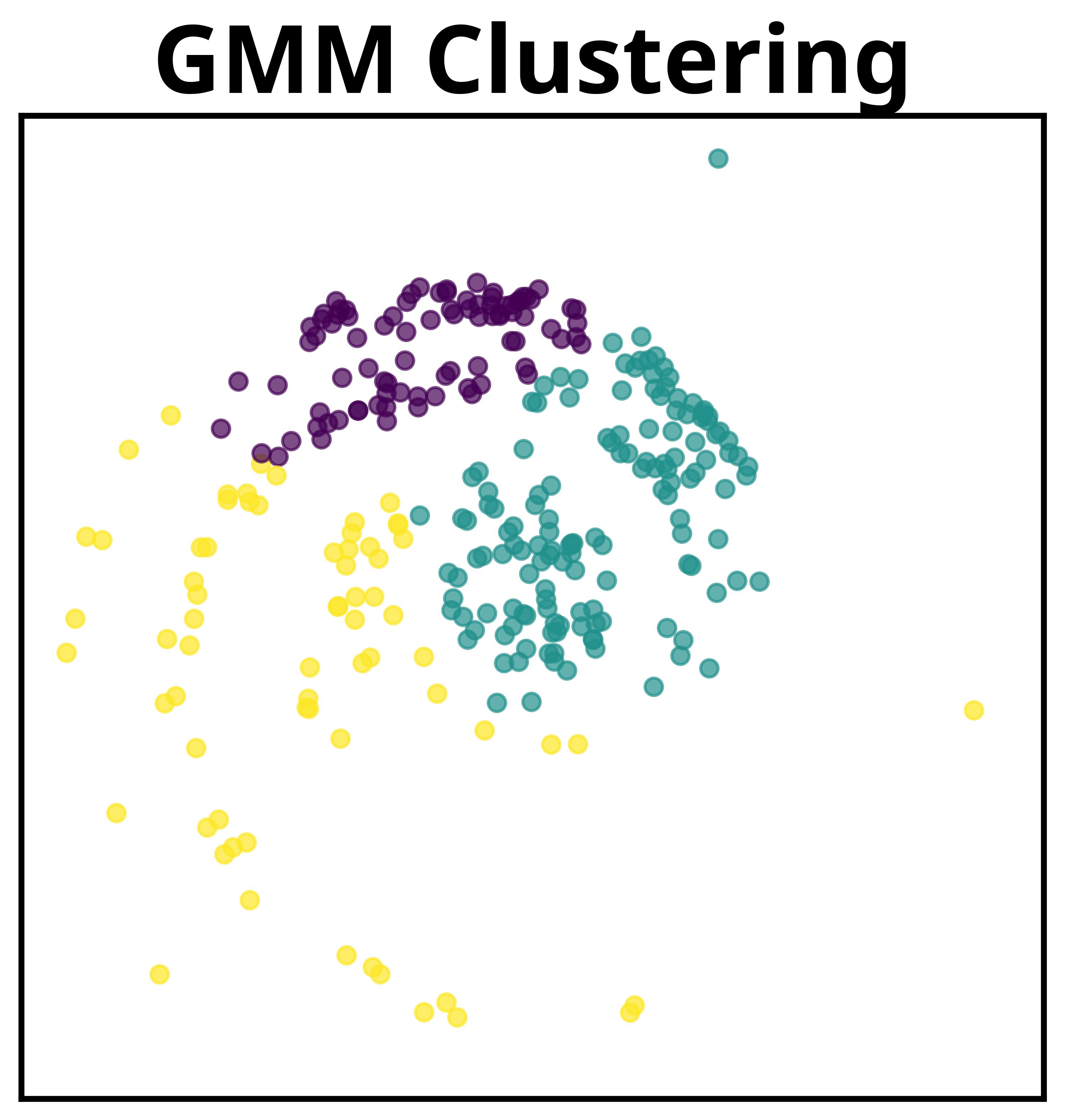}
        };
        \node (right) at (0.49\textwidth,0) {
        \includegraphics[width=0.46\textwidth]{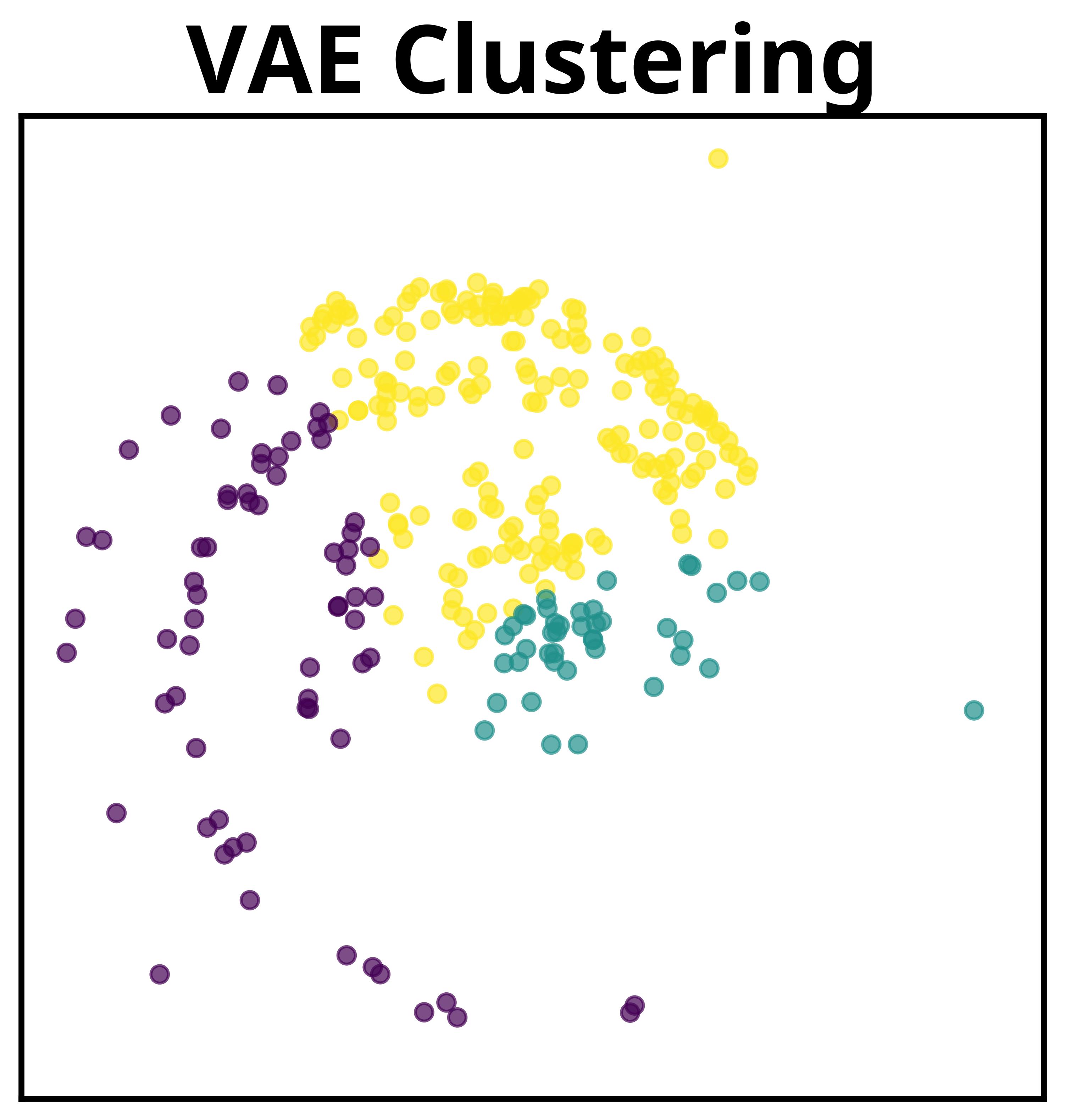}
        };
        \end{tikzpicture}
        \caption{}
        \label{fig:toy_gmm}
    \end{subfigure}
    

    \caption{In our example, where the observed data is generated by applying a spiral-shaped deformation to the underlying latent representation (Fig.~\ref{fig:toy_true}, left), MPM successfully learns the non-linear transformation $T$ using multiple datasets. The resulting cluster assignments (\ref{fig:toy_mpm}, right) more closely match the ground truth (\ref{fig:toy_true}, right) compared to those from a canonical GMM or VAE (\ref{fig:toy_gmm}).}
    \label{fig:toy}
\end{figure*}

We consider settings with a collection of $M$ related datasets $\{\mathcal{D}_i\}_{i=1}^M$, where each dataset $\mathcal{D}_i = \{x_{ij}\}_{j=1}^{N_i}$ consists of (potentially high-dimensional) observations from an underlying latent variable model. To model this structure, we introduce latent \textit{local parameters} $z_{ij}$ for individual datapoints, \textit{dataset parameters} $\Lambda = \{\lambda_i\}_{i=1}^M$ that encode variation across datasets, and \textit{global parameters} $\eta,\theta$ that govern the generative process of the dataset parameters and observations, respectively. This hierarchical setup differs from most previous work, which do not specify separate levels for global and dataset parameters~\citep{Krishnan2017StructuredIN,Johnson2016ComposingGM}. A graphical model representation of this framework is depicted in Figure~\ref{fig:mpm_graph}. 

The objective is to learn the global and dataset parameters that maximize the data likelihood: 
\begin{align*}
    \mathcal{L}(\Lambda, \theta, \eta) &\coloneqq \sum_i \mathcal{L}_i(\Lambda, \theta, \eta), \\
    \mathcal{L}_i(\Lambda, \theta, \eta) &\coloneqq \log p_\eta (\lambda_i) + \sum_j \log p_\theta(x_{ij} \mid \lambda_i).
\end{align*}

Like most latent variable models, the true posterior over $z_{ij}$ is generally intractable. One standard solution to this problem uses variational inference, which posits an approximate posterior $q$ and maximizes the Evidence Lower Bound (ELBO) $\mathcal{L}^{\text{ELBO}\vphantom{\left| \widehat{T} \right|}} \leq \mathcal{L}$, where $\mathcal{L}^{\text{ELBO}\vphantom{\left| \widehat{T} \right|}} \coloneqq \sum_i \mathcal{L}^\text{ELBO}_i$, and 
\vspace{-11pt}

{\small \begin{align*}
    &\mathcal{L}^\text{ELBO}_i \coloneqq \log p_\eta (\lambda_i) + \phantom{.} 
    \sum_j \mathbb{E}_q \left[ \log \frac{p_\theta(x_{ij}, z_{ij} \mid \lambda_i)}{q(z_{ij})} \right] \\
    &= \log p_\eta (\lambda_i) + \sum_j \, \mathbb{E}_q \left[ \log \frac{p_\theta(x_{ij} \mid z_{ij}, \lambda_i) p(z_{ij} \mid \lambda_i)}{q(z_{ij})} \right].
\end{align*}}


Here, $\mathcal{L}^{\text{ELBO}}$ and $\mathcal{L}^{\text{ELBO}}_i$ are implicitly understood to be functions of $\Lambda$, $\theta$, $\eta$, and $q$.

\subsection{Spiral GMM example}
To illustrate the motivation behind MPM, consider a simple toy example involving Gaussian Mixture Models (GMMs). Suppose a practitioner wants to cluster data that originates from a mixture of Gaussians, but the observations have been transformed by an unknown mapping \mbox{$x \mapsto Tx$} that potentially distorts the underlying cluster geometry. 

In our example, we choose $T$ to have the form \mbox{$(r, \theta) \mapsto (r, \theta + \alpha r + \beta)$} in polar coordinates. This transformation produces the spiral-shaped dataset depicted in Figure~\ref{fig:toy_true}, where each cluster corresponds to an arm or the center of the spiral. The parameters $\alpha$ and $\beta$ control the deformation strength and global angular offset, respectively.

Inferring the transformation $T$ from a single dataset is under-specified. However, when multiple datasets are available and each transformed by the same mapping, MPM can exploit shared structure to learn a coherent latent representation by estimating the mapping $\widehat{T}$. As illustrated in Figure~\ref{fig:toy_mpm}, MPM recovers the latent space up to the global rotation $\beta$, yielding cluster assignments that match the ground truth. In contrast, a standard GMM in Figure~\ref{fig:toy_gmm} fails to recover the original cluster structure.

Note that learning the transformation $T$ is equivalent to learning an induced distance function on the data. More generally, when a specific parameterization of $T$ is unavailable, we can instead model it using an expressive function class, such as a neural network. In such cases, we demonstrate that the model remains tractable by optimizing an appropriate surrogate objective.

\subsection{Fast and scalable inference using surrogate objectives}

In classical PGMs, optimization of $\mathcal{L}^\text{ELBO}_i$ is typically carried out using coordinate-ascent procedures such as variational EM. However, this approach for MPM poses two computational challenges: (i) the number of parameters grows linearly with the number of datasets, since each dataset introduces $\lambda_i$ and $z_{ij}$, and (ii) optimizing $q$ becomes computationally expensive for complex models, because it requires repeated evaluation of $p_\theta(z_{ij} \mid x_{ij}, \lambda_i)$. 

To address these issues, we define $q$ and $\Lambda$ implicitly as functions of the global parameters $\theta$ and $\eta$, by treating them as local partial optimizers of the ELBO. Solving this optimization is still intractable in general, so we instead introduce a tractable surrogate objective $\widehat{\mathcal{L}}^\text{ELBO}_i$ that admits efficient updates of $q$ and $\Lambda$. The surrogate objective takes the form:
\begin{align*}
    \widehat{\mathcal{L}}^\text{ELBO}_i &(\Lambda, \phi, \eta, q) \coloneqq \log p(\lambda_i \mid \eta) + \phantom{.} \\
    &\sum_j \mathbb{E}_q \left[ \log \frac{\exp\{\psi_\phi(z_{ij} \mid x_{ij}, \lambda_i)\} p(z_{ij} \mid \lambda_i)}{q(z_{ij})} \right], 
\end{align*} 
Here, $\psi_\phi$ is a surrogate potential from a recognition network parameterized by $\phi$, analogous to the inference network in variational autoencoders (VAEs). 

\begin{algorithm}[t]
\caption{Training and inference procedure for meta-probabilistic modeling}
\setlength{\baselineskip}{1.1\baselineskip}
\begin{algorithmic}[1]
\REQUIRE Datasets $\{\mathcal{D}_i\}_{i=1}^M$, inner optimization steps $T$, minibatch size $B$, learning rate $\alpha$
\ENSURE Parameters $\lambda^0, \theta, \phi, \eta$
\STATE \textbf{Initialize} $\vartheta = \{\lambda^0, \theta, \phi, \eta\}$
\WHILE{not converged}
  \STATE Sample minibatch $\mathcal{B} \subseteq \{\mathcal{D}_i\}^M_{i=1}$ with $|\mathcal{B}| = B$
  \STATE \textbf{Initialize} $\Lambda^{(0)}_{\phi, \eta} \gets \{\lambda^0\}^M_{i=1}$
    \FOR{$t = 1$ \TO $T$}
      \STATE $q^{(t)}_{\phi, \eta} \leftarrow \arg\max_{q} \ \widehat{\mathcal{L}}^\text{ELBO}_\mathcal{B} \left(\Lambda_{\phi, \eta}^{(t-1)},\phi,\eta,q \right)$ 
      \STATE $\Lambda^{(t)}_{\phi, \eta} \leftarrow \arg\max_{\Lambda} \ \widehat{\mathcal{L}}^\text{ELBO}_\mathcal{B} \left( \Lambda,\phi,\eta,q^{(t)}_{\phi, \eta} \right)$ 
    \ENDFOR
    \STATE $\mathcal{L}^\text{MPM}_\mathcal{B} \gets \mathcal{L}^\text{ELBO}_\mathcal{B} \left( \Lambda^{(T)}_{\phi, \eta},\theta,\eta,q^{(T)}_{\phi, \eta} \right)$
    \STATE $\vartheta \gets \texttt{SGD}(\vartheta, \nabla_\vartheta \mathcal{L}^\text{MPM}_\mathcal{B}, \alpha)$
\ENDWHILE
\STATE \textbf{return} $\lambda^0, \theta, \phi, \eta$
\end{algorithmic}
\label{alg:training}
\end{algorithm}

This construction is inspired by~\cite{Johnson2016ComposingGM}, which exploit conjugacy to obtain closed-form updates for $q$. For MPM, we instead choose $\psi_\phi$ such that $q$ and $\Lambda$ are jointly optimizable via a coordinate ascent procedure. 

To make the dependence on the global parameters explicit, we define the following iterative updates:
\begin{align*}
    \Lambda^{(t+1)}_{\phi, \eta} &= \arg\max_{\Lambda} \widehat{\mathcal{L}}^\text{ELBO}_i (\Lambda, \phi, \eta, q^{(t)}_{\phi, \eta} ), \\
    q^{(t+1)}_{\phi, \eta} &= \arg \max_q \widehat{\mathcal{L}}^\text{ELBO}_i (\Lambda^{(t+1)}_{\phi, \eta}, \phi, \eta, q),
\end{align*}
with a shared initialization $\Lambda^{(0)}_{\phi, \eta} = \{\lambda^0\}_{i=1}^M$. The meta-probabilistic modeling objective is defined as
\[ \mathcal{L}^\text{MPM}(\lambda^0, \theta, \phi, \eta) \coloneqq \sum_i \mathcal{L}_i^\text{ELBO} (\Lambda^T_{\phi, \eta}, \theta, \eta, q^T_{\phi, \eta}). \]
The MPM objective provides a lower bound on the data likelihood, in the sense that
\begin{align*}
    \mathcal{L}(\Lambda, \theta, \eta) &\geq \max_{\Lambda, q} \sum_i \mathcal{L}^\text{ELBO}_i (\Lambda, \theta, \eta, q) \\
    &\geq \sum_i \mathcal{L}_i^\text{ELBO} (\Lambda^T_{\phi, \eta}, \theta, \eta, q^T_{\phi, \eta}) = \mathcal{L}^\text{MPM}.
\end{align*}
The model is trained using a bi-level optimization procedure outlined in Algorithm~\ref{alg:training}. In the inner loop (lines 5-8), we optimize over $q$ and $\Lambda$, while holding the generative model and recognition network fixed. The outer meta-learning step (lines 9-10) then updates the global parameters $\theta, \eta$, recognition network parameters $\phi$, and learnable initialization $\lambda^0$. 

To ensure tractable inference, the surrogate potential $\psi_\phi$ must be chosen so that the inner optimization can be done efficiently. In such cases, optimizing over $\phi$ is effectively learning a recognition network that best approximates the posterior under the chosen $\psi_\phi$. This approximation serves to make learning dataset variables computationally feasible in terms of time and parameter count. 

Moreover, we emphasize that our proposed learning algorithm is not intrinsic to the MPM methodology itself. While we present a simple and scalable optimization procedure, alternative inference approaches, such as MCMC, could also be used.

\section{TWO MPM CASE STUDIES} \label{sec:case_studies}

Our method requires access to multiple related datasets in order to learn both global and dataset-specific parameters. In this section, we consider two concrete and practical settings in which this assumption naturally arises, focusing on object-centric learning and sequential text modeling.

\subsection{MPM for object-centric learning} 

We first apply MPM to object-centric learning, which aims to identify coherent objects within an image. We cast this task as a clustering problem in which pixels are grouped according to their semantic roles. This setting is well suited for MPM because global parameters model visual generative components shared across images, while dataset and local parameters capture the composition and arrangement of objects within each individual image.

Formally, we treat each image as a dataset $\mathcal{D}_i$, where $\{x_{ij}\}_{j=1}^{N_i}$ represents its pixels. Because the goal of object-centric learning is to cluster pixels within each image into semantically coherent objects, we accordingly model the dataset parameters $\lambda_i = \{\mu_{ik}\}_{k=1}^K$ as the $K$ cluster centers of an image-specific GMM and $z_{ij}$ as the cluster assignments. The centers are themselves generated from a global GMM with centers $\eta = \{\nu_\ell\}_{\ell=1}^L$. For simplicity, we assume isotropic Gaussian components with identity covariance and uniform mixing weights. 

Under this formulation, our method uses a \emph{mixture generative model} in which the local cluster centers are mapped through a learned transformation $f_\theta$ specified by global parameters.
\begin{align*}
    p_\theta(x_{ij} \mid z_{ij} = k, \, \lambda_i) \propto \exp \left( -\frac{1}{2} \| x_{ij} - f_\theta(\mu_{ik})_j \|^2 \right).
\end{align*}
Here, $f_\theta$ can be viewed as a neural network that induces a learned distance function between the pixels $x_{ij}$ and cluster centers $\mu_{ik}$. 

We define the surrogate potential to be,
\[ \psi_\phi (z_{ij} = k \mid x_{ij}, \lambda_i) = -\frac{1}{2} \| \mu_{ik} - g_\phi(x_{ij}) \|^2. \]
Returning to the VAE analogy, $g_\phi$ is a recognition network that maps each pixel into a shared latent space, while $f_\theta$ acts as a deterministic image decoder. 

Additionally, we also consider an \emph{additive generative model} with the form
{\small \[ p_\theta(x_{ij} \mid \lambda_i) \propto \exp \left( \mkern-0.4mu -\frac{1}{2} \Big\| x_{ij} - \sum_k w(\lambda_i)_{jk} \cdot f_\theta (\mu_{ik})_j \Big\|^2 \right), \]
}where $w$ are alpha masks, normalized across clusters for each pixel. This additive decoder is commonly used in object-centric learning, in part due to its favorable theoretical identifiability properties \citep{pmlr-v97-greff19a,lachapelle2023additivedecoderslatentvariables}. We also consider this model in order to enable direct comparison with existing object-centric learning methods \citep{Locatello2020ObjectCentricLearning,Wang2023SlotVAEOS}.

Intuitively, both models fit an image-specific GMM in a shared latent space, with a prior over the cluster centers. This design admits closed-form update steps for $q$ and $\Lambda$ using coordinate ascent.

\begin{proposition} \label{prop:opt}
    For fixed $\phi$ and $\eta$, the following updates for $q$ and $\Lambda$ (Algorithm~\ref{alg:training}, lines 6-7) do not decrease the surrogate ELBO.
    \begin{gather*}
        q(z_{ij} = k) \propto \exp\left( -\frac{1}{2} \| \mu_{ik} - g_\phi(x_{ij}) \|^2 \right), \\
        \mu_{ik} = \frac{\sum_{\ell} r_{ik\ell} \nu_\ell + \sum_j s_{ijk} g_\phi(x_{ij})}{\sum_{\ell} r_{ik\ell} + \sum_j s_{ijk}},
    \end{gather*}
    where,
    \begin{align*}
        r_{ik\ell} &= \frac{\exp\left( -\frac{1}{2} \|\mu_{ik} - \nu_\ell \|^2 \right)}{\sum_{\tilde \ell} \exp\left( -\frac{1}{2} \|\mu_{ik} - \nu_{\tilde \ell} \|^2 \right)}, \\
        s_{ijk} &= \frac{\exp \left(-\frac{1}{2} \| \mu_{ik} - g_\phi(x_{ij}) \|^2 \right)}{\sum_{\tilde k} \exp \left(-\frac{1}{2} \| \mu_{i{\tilde k}} - g_\phi(x_{ij}) \|^2 \right)}.
    \end{align*}
\end{proposition}
The proof is provided in Appendix~\ref{app:opt}. Since these updates can be computed efficiently, the optimization is tractable and can be carried out using Algorithm~\ref{alg:training}.

\textbf{Connection with Slot Attention.} Many modern architectures for object-centric learning build on the Slot Attention module introduced by \cite{Locatello2020ObjectCentricLearning}. We show that our proposed approach for this task is closely related to Slot Attention and can be viewed through a similar modeling lens.

At a high level, the Slot Attention model encodes each image to a latent representation $z$, and a set of $K$ slots $s$ is iteratively refined using an attention mechanism. Each slot is intended to represent a distinct object in the scene. A decoder then maps each slot to an object-specific representation, which are then combined additively.

The primary contribution of Slot Attention is the slot refinement algorithm, which iteratively updates the set of slots by computing scaled dot-product attention between the latent representation and the slots. Let $W_q$, $W_k$, and $W_v$ be the query, key, and value projection matrices, respectively, and let $s^{(t)}$ represent the slot embeddings at iteration $t$. The update at each step is given by:
\begin{align*}
    A^{(t)} &= \mathsf{Softmax} \left( \frac{(W_k z) (W_q  s^{(t-1)})^T}{\sqrt{D}}, \, \mathsf{axis=slots} \right), \\
    u^{(t)} &= \mathsf{WeightedMean}(\mathsf{weights=}A^{(t)}, \mathsf{values=}W_v z), \\
    s^{(t)} &= \mathsf{SlotUpdate}(u^{(t)}, s^{(t-1)}).
\end{align*}
The initial slots $s^{(0)}$ are sampled from a learned Gaussian distribution and iteratively refined over $T$ rounds. Additional details of the Slot Attention algorithm are provided in Appendix~\ref{app:slot_attn_details}.

Slot Attention shares several structural similarities with our meta-probabilistic model for object-centric learning. In particular, the dataset-level parameters $\Lambda$ correspond to the slots, while the global parameters $\theta$ define the decoder. When viewed through this lens, the slot refinement procedure closely resembles the inner optimization steps of Algorithm~\ref{alg:training} (lines 6–7). The attention weights $W^{(t)}$ correspond to the approximate posterior $q$, with similarity measured using a scaled dot-product attention instead of Euclidean distance, while the weighted mean update $u^{(t)}$ is a special case when $r_{ik\ell} = 0$.

Hence, Slot Attention admits a precise probabilistic interpretation by treating the iterative updates as approximate likelihood maximization in a latent clustering model. From this perspective, the effectiveness of Slot Attention primarily arises from its implicit role as a meta-probabilistic model, rather than from the attention mechanism itself. 

The connection to MPM also provides a principled foundation for extending Slot Attention. In particular, we have considered a setting in which the slots themselves are generated by a global GMM parameterized by $\eta$. This extension enables the model to learn object-centric representations while simultaneously clustering the objects themselves to discover shared features across datasets.

\subsection{MPM for sequential text modeling}

We extend the idea of clustering pixels in images to text by grouping words within a document to uncover underlying semantic and syntactic themes. In this setting, we consider a corpus of documents, where each document corresponds to a dataset $\mathcal{D}_i$. The observations $x_{ij} \in \{1, \ldots, V\}$ are the words in document $i$, where $V$ denotes vocabulary size. The dataset parameters $\lambda_i = \{\mu_{ik}\}_{k=1}^K$ represent $K$ latent topic embeddings. 

We posit the following generative model:
\begin{align*}
    p(z_{ij} = k \mid \lambda_i) &= 1/K, \\
    p(x_{ij} = v \mid z_{ij} = k, \lambda_i) &= \mathsf{Softmax}(f_\theta(\mu_{ik}, s_j))_v
\end{align*}
where $f_\theta$ maps a topic embedding $\mu_{ik}$ together with a positional encoding $s_j$ to a distribution over words. We define the surrogate potential
\[ \psi_\phi(z_{ij} = k \mid x_{ij}, \lambda_i) = -\frac{1}{2} \|\mu_{ik} - g_\phi(x_{ij}) \|^2, \]
where $g_\phi$ is a recognition network that produces contextual embeddings for each word in \mbox{document $i$}. Because we reuse the surrogate potential from the object-centric learning setting, the resulting updates for $q$ and $\Lambda$ are identical. Our formulation corresponds to fitting a document-specific GMM in a shared latent space, with each mixture component representing a latent topic or theme.

\textit{Remark.} So far, we have presented applications of MPM while keeping the models $f_\theta$ and $g_\phi$ abstract, imposing no restrictions on their specific forms. Our methodology is agnostic to the choice of these components and allows for any suitable high-capacity parameterization appropriate to the domain. For example, in object-centric learning we use standard CNN architectures, whereas in the text modeling setting, we use a pre-trained language model to obtain contextual word embeddings. Additional details on the model architecture are provided in Appendix~\ref{app:training}.

\section{RELATED WORK} \label{sec:related_work}

Several lines of prior research connect PGMs, deep learning, and meta-learning. We review the most relevant directions and studies below.

\textbf{Probabilistic~graphical~models.}~Various PGMs (e.g., Bayesian Networks ~\citep{Pearl1986FusionPA}, Markov Random Fields ~\citep{Boykov1998MarkovRF}, latent variable models) have been proposed across diverse domains, such as medical diagnosis~\citep{McLachlan2020BayesianNI}, sensing~\citep{Diebel2005AnAO}, and natural language processing~\citep{Blei2003LatentDA}. However, these models demand careful specification, which can be difficult in heterogeneous or high-dimensional data.

Deep generative architectures such as VAEs~\citep{Kingma2013AutoEncodingVB} and Deep Boltzmann Machines~\citep{Salakhutdinov2009DeepBM,Goan2020BayesianNN} alleviate some of these challenges by using neural networks to approximate complex conditional distributions. This improves generative capabilities, but often at the expense of the well-defined latent semantics that make PGMs interpretable~\citep{Svensson2019InterpretableFM}.

\textbf{Hybrid~deep-probabilistic~models.}~Several studies have combined neural networks with the structured reasoning of PGMs. For example, Deep Poisson Factor Analysis~\citep{Gan2015ScalableDP} and Deep Latent Dirichlet Allocation~\citep{Cong2017DeepLD} replace classical priors and likelihoods with neural parameterizations. However, these approaches are often model-specific and rely heavily on sampling-based inference, which limits their generality and scalability. In contrast, MPM provides a general approach for learning generative models for a broad class of latent variable models. 

\textbf{Structured~variational~inference.}~A similar line of work explores combining probabilistic structure with neural inference using variational methods. Structured VAEs~\citep{Johnson2016ComposingGM} augment graphical models with neural components for structured latent representations. However, their framework learns generative mechanisms from a single dataset, which does not capture cross-data structure. \citet{Krishnan2017StructuredIN} integrate VAEs with continuous state-space models, using inference networks to model temporal latent structure. While effective for nonlinear dynamical systems, the approach does not readily extend to more general model classes. 

Our method extends structured variational inference in two key aspects: (1) it applies to a broad class of tractable latent variable models, and (2) it explicitly separates global generative structure from dataset-specific variation. We show this approach demonstrates connections to existing models and meta-learning, and is empirically able to discover latent structure within and across complex datasets.

\begin{table}[t]
    \renewcommand{\arraystretch}{1.2}
    \centering
    \caption{
    On the Tetrominoes dataset, the additive meta-probabilistic model achieves the highest mean ARI score across five runs with different random seeds. In contrast, existing models (i.e. Slot Attention and Slot VAE) exhibit significant variability in performance across runs, while generative baselines are not well suited for the object-centric learning task. Error bars indicate one standard deviation.}
    \label{tab:ari}
    \begin{tabular}{cc}
        \toprule
        Model & ARI (\%) \\
        \midrule
        GMM & $77.38 \pm 0.45\phantom{0}$ \\
        VAE & $16.81 \pm 9.36\phantom{0}$ \\
        Latent diffusion & $14.84 \pm 7.41\phantom{0}$ \\
        Slot attention & $85.40 \pm 13.86$ \\
        Slot VAE & $25.35 \pm 34.57$ \\
        Mixture MPM (Ours) & $52.93 \pm 3.20\phantom{0}$ \\
        Additive MPM (Ours) & \underline{$\mathbf{96.46 \pm 1.89}\phantom{0}$} \\
        \bottomrule
    \end{tabular}
\end{table}

\textbf{Meta-learning~and~probabilistic~models.}~Our approach also connects to meta-learning, which is a paradigm that aims to generalize across tasks or datasets~\citep{Hospedales2022MetaLearningIN}. Extensions of meta-learning to probabilistic models include meta-amortized inference~\citep{Edwards2016TowardsAN}, where a dataset-specific context governs the latent space, and probabilistic task models~\citep{Nguyen2021ProbabilisticTM}, which combine a VAE with a Gaussian LDA prior for discovering task-themes. While we share a high-level structure with these works, our goal is to provide a more general framework for combining latent variable models with neural architectures. We develop a scalable learning algorithm for this class of models, which provides a probabilistic understanding of the existing Slot Attention architecture.

Other work has linked meta-learning to Bayesian inference. For instance,~\citet{Grant2018RecastingGB} show that Model-Agnostic Meta-Learning~\citep{pmlr-v70-finn17a} can be interpreted as hierarchical Bayesian inference. While we also adopt a hierarchical Bayesian perspective, our framework differs by explicitly separating global components from local parameters. This enables learning the underlying generative process itself across datasets while still allowing flexible adaptation to dataset-level variation.

\section{EXPERIMENTS} \label{sec:exp}

We evaluate our proposed method on object-centric learning and sequential text modeling tasks. 
The experiments demonstrate that MPM (1) discovers shared generative mechanisms, (2) captures dataset-specific latent variables that form semantically meaningful clusters, and (3) identifies high-level latent attributes within each group. All code for our model and experiments are available at: \href{https://github.com/kzhangm02/mpm}{\texttt{https://github.com/kzhangm02/mpm}}.

\textbf{Datasets.} For object-centric learning, we use the Tetrominoes dataset~\citep{multiobjectdatasets19}, which comprises 10,000 images of three non-overlapping shapes. Each shape varies in position, color, and type, selected from a fixed set of tetrominoes. For the text modeling experiments, we use a subset of the AP News corpus~\citep{Harman1993OverviewTREC}, consisting of approximately 2,200 news articles from the Associated Press. In both domains, we partition the data into 80\% training, 10\% validation, and 10\% test splits.

\begin{figure}[t]
    \centering
    \includegraphics[width=\linewidth]{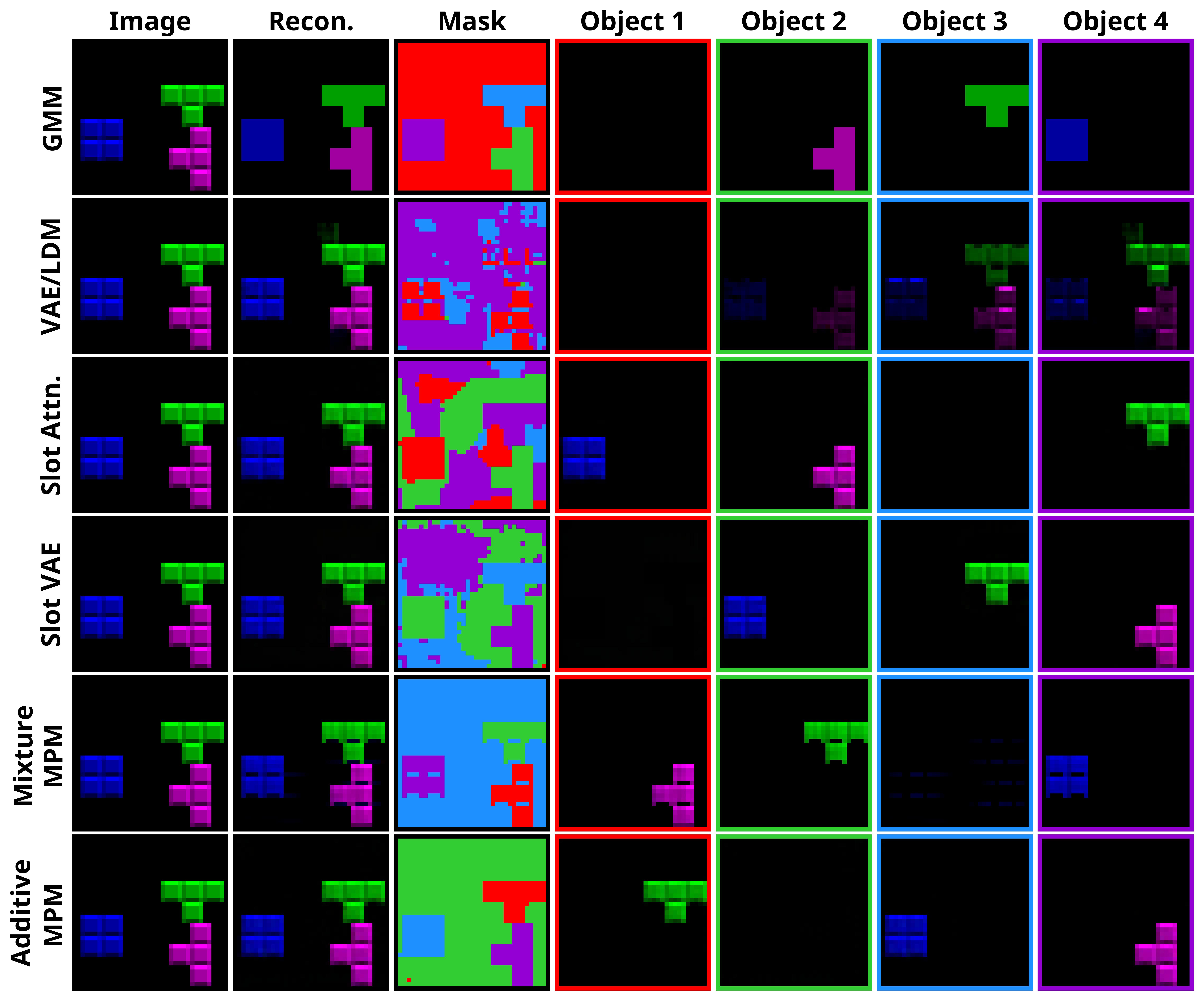}
    \caption{MPM produces high-quality images and more precise object segmentation masks compared to existing architectures. Border colors correspond to the alpha mask colors shown in the third column.}
    \label{fig:objects}
\end{figure}

\begin{figure*}[ht!]
    \centering
    \begin{subfigure}[t]{0.48\textwidth}
        \centering
        \includegraphics[width=\linewidth]{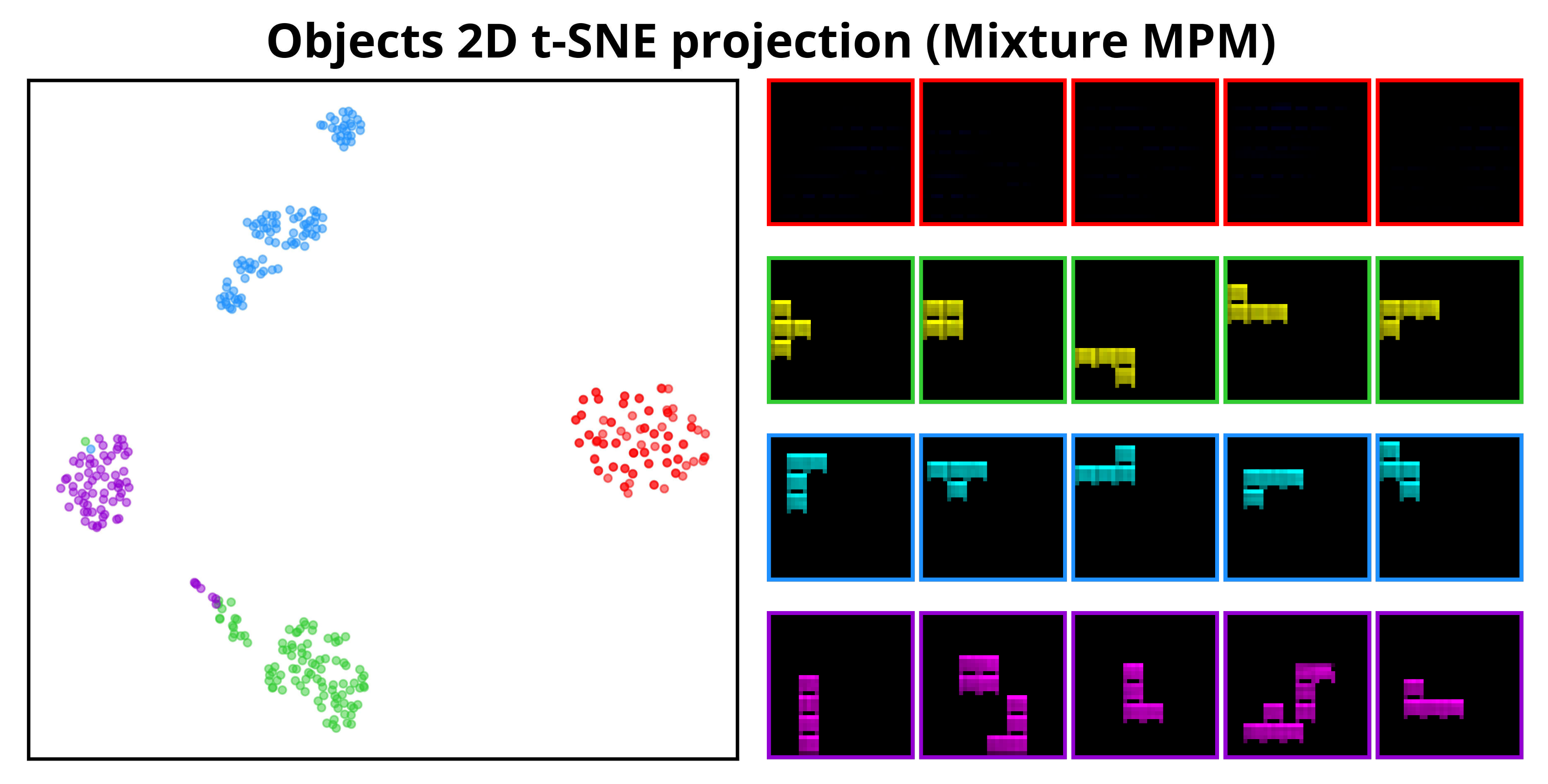}
        \caption{}
        \label{fig:global_clusters_mix}
    \end{subfigure}
    \hfill
    \begin{subfigure}[t]{0.48\textwidth}
        \centering
        \includegraphics[width=\linewidth]{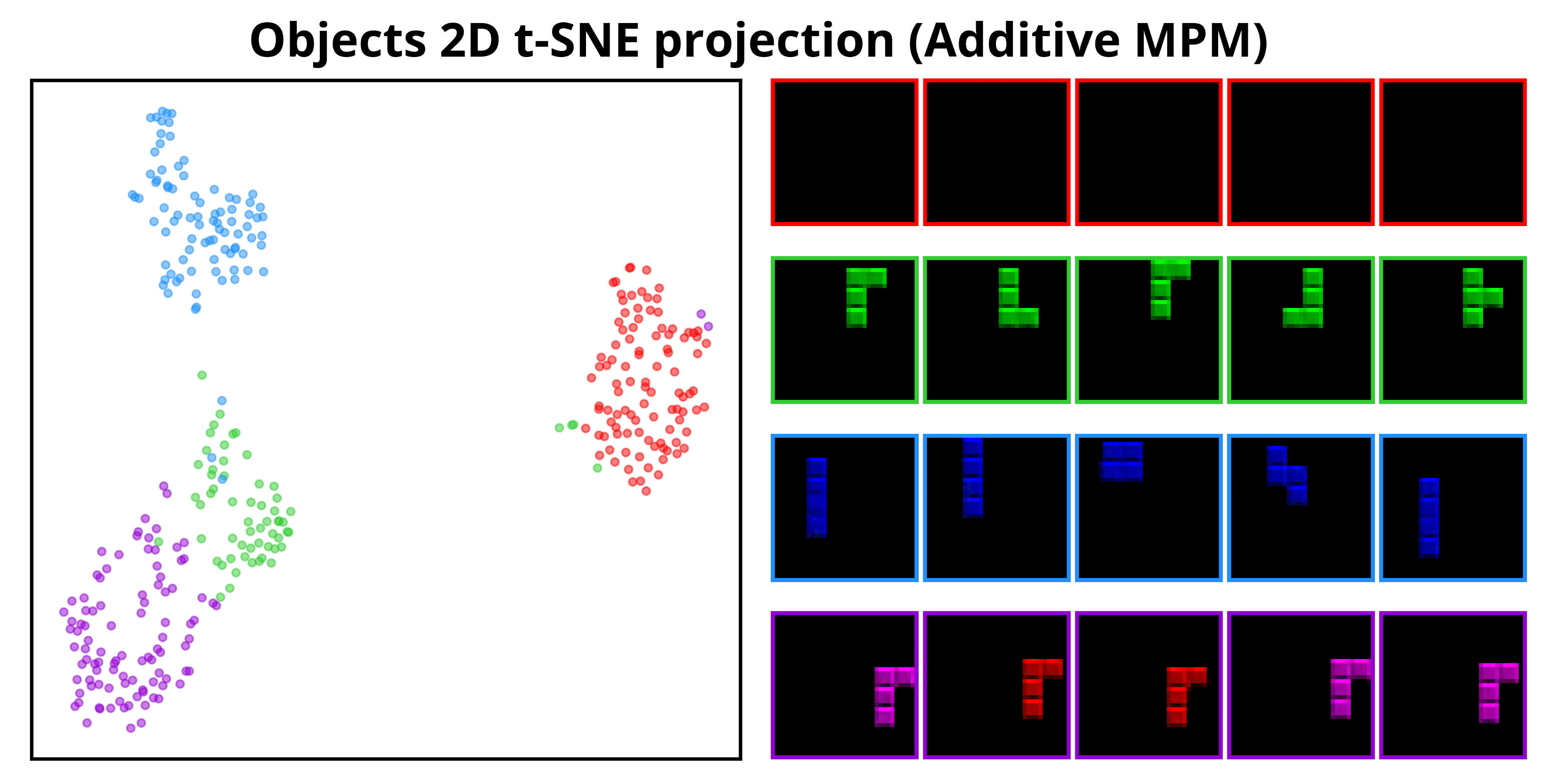}
        \caption{}
        \label{fig:global_clusters_add}
    \end{subfigure}
    \caption{Our meta-probabilistic model identifies global object clusters that align with latent attributes such as color, position, and shape, under both the mixture (\ref{fig:global_clusters_mix}) and additive (\ref{fig:global_clusters_add}) formulations. The left panel shows the two-dimensional t-SNE embedding, and the right panel displays the five objects with the highest responsibility scores for each cluster. Border colors denote the corresponding cluster assignments.}
    \label{fig:global_clusters}
\end{figure*}

\subsection{Training and Evaluation} 

All experiments follow the training procedure in Algorithm~\ref{alg:training}. In practice, we scale the entropy regularization by a multiplicative factor $\beta < 1$, similar to the $\beta$-VAE~\citep{Higgins2016betaVAE}, as this modification yields improved empirical performance. We analyze the effect of the regularization in Appendix~\ref{app:beta}.

For object-centric learning, we set $K = 4$ dataset-level clusters corresponding to the three foreground objects and the background, $L = 100$ global object clusters, and $\beta = 0.01$. The model uses the standard convolutional neural network (CNN) architecture. The sequential text modeling experiment uses $K = 5$ dataset-level clusters, $L = 100$ global topic clusters, and $\beta = 0.1$. To obtain contextualized word embeddings for the recognition model $g_\phi$, we extract token embeddings from a pretrained BERT model and average the subword representations. Additional training details are provided in Appendix~\ref{app:training}.

Following prior object-centric learning work, we evaluate performance using qualitative visualizations and the Adjusted Rand Index (ARI). ARI measures clustering agreement by comparing pairwise assignments. Zero corresponds to random clustering, while one indicates perfect agreement.

For sequential text modeling, we report the maximum and mean UMass topic coherence scores across all topics on the test set. Higher coherence values indicate stronger semantic consistency. We assess interpretability by examining the most representative words for each topic. These terms are selected using the term frequency–inverse document frequency (tf-idf), which identifies distinctive words in a topic.

\subsection{Object-centric learning results}

Table~\ref{tab:ari} reports ARI scores for our meta-probabilistic model using both mixture and additive formulations. We compare against classical clustering baselines (GMMs), generative models including a Variational Autoencoder (VAE) and a Latent Diffusion Model (LDM), and established object-centric architectures such as Slot Attention~\citep{Locatello2020ObjectCentricLearning} and Slot VAE~\citep{Wang2023SlotVAEOS}.

The additive model achieves the best performance across five training runs. Standard VAEs and LDMs obtain low ARI scores, as their objectives focus on generation rather than structured decomposition. While Slot Attention and object-centric VAEs learn object-level representations, they exhibit significant training instability across random seeds, consistent with~\citet{Locatello2020ObjectCentricLearning}. The mixture variant performs worse than the additive model. Empirically, this is due to a tendency to merge objects with similar color or shape into a single slot.

We visualize the learned structural representations in Figure~\ref{fig:objects}. The GMM baseline groups pixels primarily by color, but fails to capture object-specific structure and fine details. In contrast, the VAE and LDM generate high-quality images yet do not separate individual objects. For brevity, we present their results jointly, as they are visually similar.

Existing object-centric models achieve high-quality and semantically meaningful reconstructions. However, the spatial segmentation is often imprecise, with background pixels assigned to object slots. By comparison, our meta-probabilistic models produce sharp spatial separation between objects and background while maintaining high reconstruction quality and exhibit minimal training instability.

MPM can also discover clusters of objects across images. To visualize these global groupings, we compute responsibility scores $r_{ik\ell}$, which measure the contribution of each global cluster $c_\ell$ to a given object. For a selected subset of clusters, Figure~\ref{fig:global_clusters} shows the five objects with the highest responsibility scores, together with their two-dimensional t-SNE embeddings~\citep{JMLR:v9:vandermaaten08a}. The results reveal that the model organizes objects according to latent attributes, such as shape, color, and position, demonstrating that the learned global structure is semantically meaningful. In contrast, most existing object-centric learning architectures cannot identify any global structure across objects.

\subsection{Sequential text modeling results}

For sequential text modeling, Table~\ref{tab:umass} reports the UMass coherence score on the test corpus, using the top 10 words from each global topic ranked by tf-idf. Unused topics are excluded from the evaluation.

\begin{table}[t]
    \renewcommand{\arraystretch}{1.28}
    \centering
    \caption{MPM obtains competitive mean and maximum UMass coherence scores across topics in the AP corpus, compared to existing neural topic modeling approaches. We report the mean and standard deviation across five runs with different random seeds.}
    \label{tab:umass}
    {\small \begin{tabular}{ccc}
        \toprule
        Model & UMass (mean) & UMass (max) \\
        \midrule
        LDA & $-0.753 \pm 0.015$ & $-0.215  \pm 0.076$\\
        NVDM & $-1.072 \pm 0.028$ & $-0.746 \pm 0.065$ \\
        GSM & $-0.500 \pm 0.049$ & $-0.224 \pm 0.026$ \\
        NTM & $-0.700 \pm 0.030$ & $-0.418 \pm 0.026$ \\
        NTMR & $-1.071 \pm 0.011$ & $-0.704 \pm 0.030$ \\
        WeTe & $-0.589 \pm 0.015$ & $-0.167 \pm 0.115$ \\
        FASTopic & \underline{$\mathbf{-0.462 \pm 0.019}$} & $-0.012 \pm 0.012$ \\
        MPM (Ours) & $-0.499 \pm 0.059$ & \underline{$\mathbf{-0.011 \pm 0.008}$} \\
        \bottomrule
    \end{tabular}}
\end{table}

Because MPM is designed for latent structure discovery across datasets, we compare against neural topic modeling approaches, which typically parameterize probabilistic models with neural networks to infer topic structure. Latent Dirichlet Allocation (LDA) acts as a classical baseline. In our evaluation, we examine the Neural Variational Document Model (NVDM)~\citep{Miao2016NeuralVI}, Gaussian Softmax Topic Model (GSM)~\citep{Miao2017DiscoveringDL}, Neural Topic Model and its regularized variant (NTM/NTM-R)~\citep{Ding2018CoherenceAwareNT}, mixture-based embedding models (WeTe)~\citep{Wang2022RepresentingMO}, and FASTopic~\citep{Wu2024FASTopicPT}.

Our method achieves competitive coherence scores relative to prior work. The mean UMass coherence is similar to using FASTopic and GSM, while the maximum score is matched only by FASTopic. We additionally evaluate coherence using Normalized Pointwise Mutual Information in Appendix~\ref{app:npmi} and find similar performance. An analysis of scaling behavior is also provided in Appendix~\ref{app:scaling}.

Figure~\ref{fig:topics} shows representative words for each topic at both the document and corpus levels. Within individual sentences, topics tend to reflect syntactic structure. For example, the green topic primarily contains punctuation, whereas the purple topic consists of prepositions and articles. We also present five example global topics, which generally exhibit greater semantic coherence, though they occasionally capture syntactic elements such as punctuation.

We emphasize that the aim of our text experiments is not to achieve state-of-the-art performance in topic modeling, but rather demonstrate practical use cases of our general probabilistic modeling methodology, which naturally extends to topic modeling.

\begin{figure}[t!]
    \centering
    \begin{subfigure}[t]{\linewidth}
        \centering
        \includegraphics[width=0.9\linewidth]{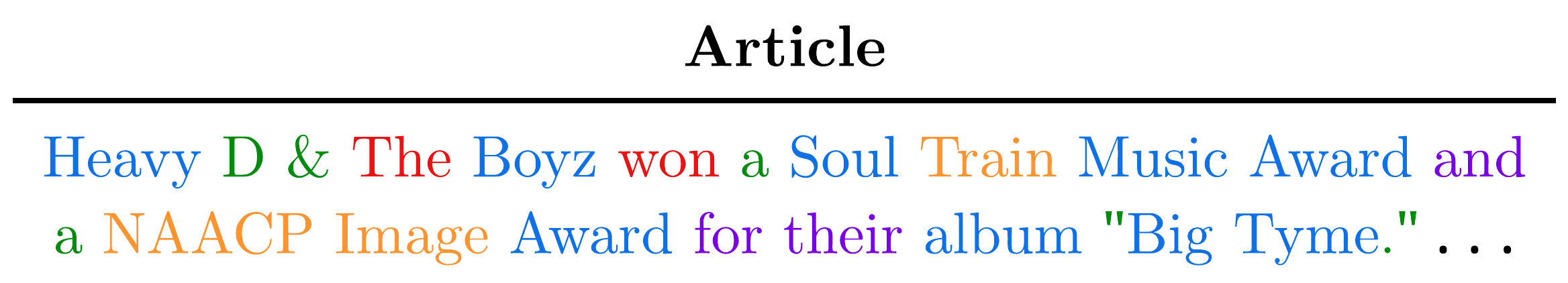}
        \caption{}
        \label{fig:example_document}
    \end{subfigure}
    \vspace{5pt}
    
    \begin{subfigure}[t]{\linewidth}
        \centering
        \includegraphics[width=0.9\linewidth]{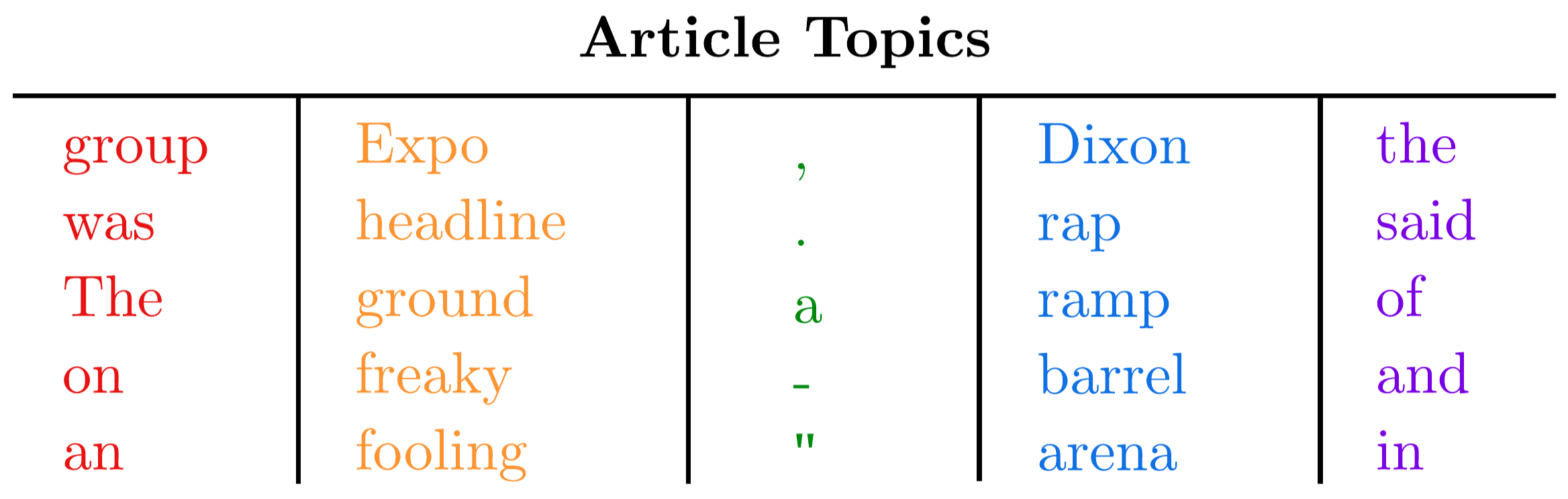}
        \caption{}
        \label{fig:topics_document}
    \end{subfigure}
    \vspace{5pt}

    \begin{subfigure}[t]{\linewidth}
        \centering
        \includegraphics[width=0.9\linewidth]{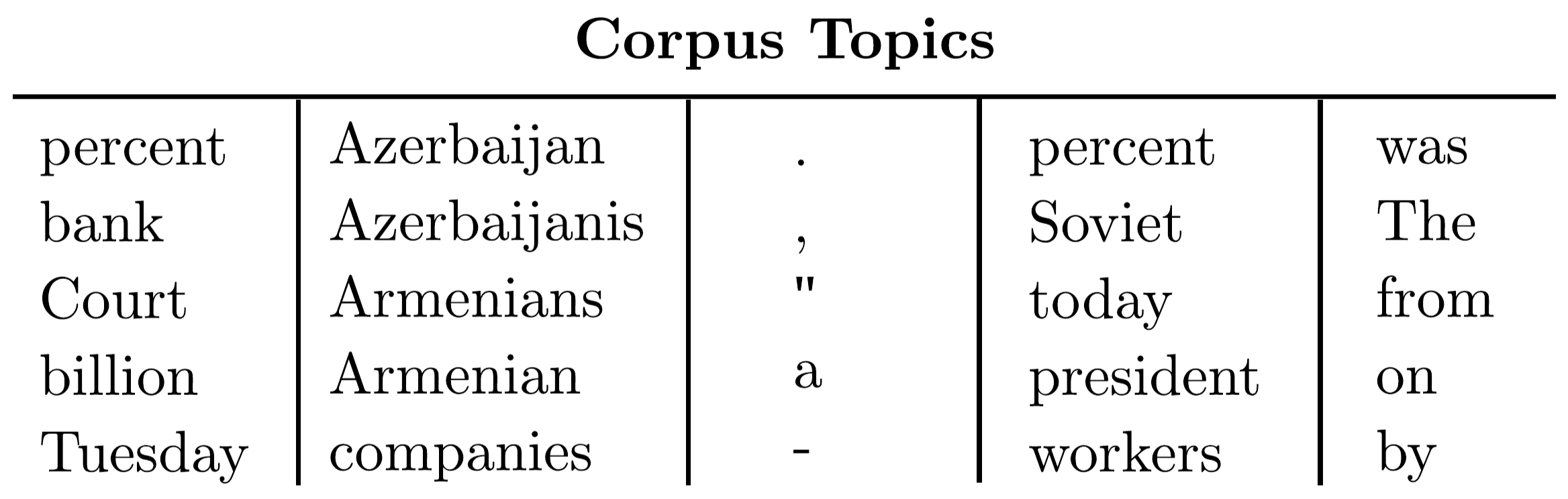}
        \caption{}
        \label{fig:topics_global}
    \end{subfigure}
    \caption{
    In the test article (\ref{fig:example_document}), document-level topics (\ref{fig:topics_document}) primarily capture syntactic structure, whereas corpus-level topics (\ref{fig:topics_global}) have more coherent semantic groupings. At each level, the top five words per topic are listed from top to bottom.}
    \label{fig:topics}
\end{figure}

\section{DISCUSSION} \label{sec:disc}
Probabilistic models are often limited by assumptions on the generative process. In this work, we propose a meta-probabilistic modeling method that learns the generative process itself from a collection of related datasets by hierarchically decomposing the generative mechanism into global and dataset-specific parameters. We develop an efficient and scalable training algorithm by deriving a tractable surrogate likelihood bound.

Our experiments show that MPM can effectively combine the expressive modeling capacity of neural networks with the interpretable structure of classical latent variable models. We also demonstrate that the Slot Attention architecture emerges as a special case of our formulation. This perspective allows us to extend the method naturally to tasks such as clustering objects across images based on latent attributes, as well as topic discovery in sequential text datasets.

\textbf{Limitations.} Meta-probabilistic modeling requires multiple related datasets to learn cross-dataset patterns. This dependence is application-specific, and not all problems will fall within this structure. Nonetheless, we provide a principled and natural way to model and learn cross-dataset relationships when they do exist, such as in object-centric modeling.

\clearpage

\section*{Acknowledgments}

We thank Kartik Ahuja for helpful discussion in forming the foundational idea of this work. YW was supported in part by funding from the Office of Naval Research under grant N00014-23-1-2590, the National Science Foundation under grant No. 2310831, No. 2428059, No. 2435696, No. 2440954, a Michigan Institute for Data Science Propelling Original Data Science (PODS) grant, Two Sigma Investments LP, and  LG Management Development Institute AI Research. Any opinions, findings, and conclusions or recommendations expressed in this material are those of the authors and do not necessarily reflect the views of the sponsors.

\bibliographystyle{plainnat}
\bibliography{ref}

\clearpage
\appendix
\thispagestyle{empty}

\onecolumn

\begin{center}
{\Large \textbf{Supplementary Materials for Meta-probabilistic Modeling}}
\vspace{0.2em}
\rule{\textwidth}{0.8pt}
\end{center}

\section{PROOF OF PROPOSITION~\ref{prop:opt}} \label{app:opt}


Recall the meta-probabilistic loss and surrogate objective:
\begin{align*}
    \mathcal{L}^\text{MPM}(\lambda^0, \theta, \phi, \eta) &\coloneqq \sum_i \mathcal{L}_i^\text{ELBO} (\Lambda^T_{\phi, \eta}, \theta, \eta, q^T_{\phi, \eta}) \\
    &\coloneqq \sum_i \left[ \log p(\lambda_i \mid \eta) + \sum_j \mathbb{E}_q \left[ \log \frac{p_\theta(x_{ij} \mid z_{ij}, \lambda_i) p(z_{ij} \mid \lambda_i)}{q(z_{ij})} \right] \right]. \\
    \widehat{\mathcal{L}}^\text{ELBO} (\Lambda, \phi, \eta, q) &\coloneqq \sum_i \widehat{\mathcal{L}}^\text{ELBO}_i (\Lambda, \phi, \eta, q) \\
    &\coloneqq \sum_i \left[ \log p(\lambda_i \mid \eta) + \sum_j \mathbb{E}_q \left[ \log \frac{\exp\{\psi_\phi(z_{ij} \mid x_{ij}, \lambda_i)\} p(z_{ij} \mid \lambda_i)}{q(z_{ij})} \right] \right],
\end{align*}
where $\psi_\phi (z_{ij} = k \mid x_{ij}, \lambda_i) = -\frac{1}{2} \| \mu_{ik} - g_\phi(x_{ij}) \|^2$.

For a fixed $i$, optimizing $\widehat{\mathcal{L}}^\text{ELBO}_i$ with respect to $q$ is equivalent to maximizing
\begin{align*}
    \sum_j \mathbb{E}_q \left[ \log \frac{\exp\left( -\frac{1}{2} \| \mu_{iz_{ij}} - g_\phi(x_{ij}) \|^2 \right)}{q(z_{ij})} \right], 
\end{align*}
which is the negative KL divergence between $q$ and the unnormalized distribution $\exp \left( -\frac{1}{2} \| \mu_{iz_{ij}} - g_\phi(x_{ij}) \|^2 \right)$. Thus, the optimal $q$ satisfies 
\[ q(z_{ij} = k) \propto \exp\left( -\frac{1}{2} \| \mu_{ik} - g_\phi(x_{ij}) \|^2 \right). \]
We find the maximizing of $\Lambda$ by setting the gradient to zero. For a fixed $\mu_{ik}$,
\begin{align*}
    \nabla_{\mu_{ik}} \widehat{\mathcal{L}}^\text{ELBO} (\Lambda, \phi, \eta, q) &= \nabla_{\mu_{ik}} \left[ \log p(\mu_{ik} \mid \eta) + \sum_j \mathbb{E}_q [\psi_\phi (z_{ij} \mid x_{ij}, \lambda_i)] \right] \\
    &= \nabla_{\mu_{ik}} \left[ \log \left( \sum_{\ell=1}^L \exp \left( -\frac{1}{2} \|\mu_{ik} - \nu_\ell \|^2 \right) \right) -\frac{1}{2} \sum_{j=1}^{N_i} \sum_{k=1}^K q(z_{ij}=k) \cdot \| \mu_{ik} - g_\phi(x_{ij}) \|^2 \right] \\
    &= - \left[ \sum_{\ell = 1}^L r_{ik\ell} (\mu_{ik} - \nu_\ell) + \sum_{j=1}^{N_i} s_{ijk} (\mu_{ik} - g_\phi(x_{ij})) \right],
\end{align*}
where 
\begin{align*}
    r_{ik\ell} = \frac{\exp\left( -\frac{1}{2} \|\mu_{ik} - \nu_\ell \|^2 \right)}{\sum_{\tilde \ell} \exp\left( -\frac{1}{2} \|\mu_{ik} - \nu_{\tilde \ell} \|^2 \right)}, \quad s_{ijk} = \frac{\exp \left(-\frac{1}{2} \| \mu_{ik} - g_\phi(x_{ij}) \|^2 \right)}{\sum_{\tilde k} \exp \left(-\frac{1}{2} \| \mu_{i{\tilde k}} - g_\phi(x_{ij}) \|^2 \right)}.
\end{align*}
Setting the gradient to zero yields the update for $\mu_{ik}$:
\[ \mu_{ik} = \frac{\sum_{\ell} r_{ik\ell} \nu_\ell + \sum_j s_{ijk} g_\phi(x_{ij})}{\sum_{\ell} r_{ik\ell} + \sum_j s_{ijk}}. \]
This provides the update steps used in the meta-probabilistic inference procedure in our two case studies. \qed

\begin{algorithm}[t]
\caption{Training procedure for Slot Attention}
\setlength{\baselineskip}{1.1\baselineskip}
\begin{algorithmic}[1]
\REQUIRE Dataset $\mathcal{D} = \{x_i\}_{i=1}^n$, inner optimization steps $T$, learning rate $\alpha$
\ENSURE Parameters $\theta, \phi, \mu, \sigma$
\STATE \textbf{Initialize} $\vartheta = \{\theta, \phi, \mu, \sigma\}$
\WHILE{not converged}
  \STATE Sample $x \in \mathsf{Uniform}(\mathcal{D})$
  \STATE Sample $\varepsilon_{k} \sim \mathcal{N}(0, I_d)$
  \STATE $z \gets \mathsf{LayerNorm}(g_\phi(x))$
  \STATE $s^{(0)} \gets \mu + \sigma \cdot \varepsilon$
    \FOR{$t = 0$ \TO $T-1$}
      \STATE $s^{(t)}_\text{prev} \gets s^{(t)}$
      \STATE $s^{(t)} \gets \mathsf{LayerNorm}(s^{(t)})$
      \STATE $ A^{(t)} \gets \mathsf{Softmax} \left( \frac{(s^{(t)} W_q) (z W_k)^\top}{\sqrt{d}}, \,\mathsf{axis=slots} \right)$
      \STATE $u^{(t)} \gets \mathsf{WeightedMean}(\mathsf{weights=}A^{(t)}, \mathsf{values=}z W_v)$
      \STATE $s^{(t+1)} \gets \mathsf{GRU}(\mathsf{state = }\, s^{(t)}_\text{prev}, \mathsf{update = }\, u^{(t)})$
      \STATE $s^{(t+1)} \gets s^{(t+1)} + \mathsf{MLP}(\mathsf{LayerNorm}(s^{(t+1)}))$
    \ENDFOR
    \STATE $\mathcal{L} \gets \|x - f_\theta(s^{(T)})\|^2$
    \STATE $\vartheta \gets \texttt{SGD}(\vartheta, \nabla_\vartheta \mathcal{L}, \alpha)$
\ENDWHILE
\STATE \textbf{return} $\theta, \phi, \mu, \sigma$
\end{algorithmic}
\label{alg:slot_attn_training}
\end{algorithm}

\section{SLOT ATTENTION DETAILS} \label{app:slot_attn_details}
We provide an overview of the Slot Attention architecture for object-centric learning. Let $\mathcal{X} \subseteq \mathbb{R}^{H \times W \times 3}$ denote the image space, and let $g_\phi : \mathcal{X} \to \mathcal{Z}$ be a deterministic encoder that maps an image to a latent representation, where $\mathcal{Z} \subseteq \mathbb{R}^{N \times d}$ consists of $N$ feature vectors in $\mathbb{R}^d$. The \emph{Slot Attention module} transforms these features into a set of $K$ slots, each represented in the slot space $\mathcal{S} \subseteq \mathbb{R}^{d}$. A deterministic decoder \mbox{$f_\theta : \mathcal{S}^K \to \mathcal{X}$} then maps the slots back to the image space. Typically, the encoder and decoder are implemented using CNNs with spatial broadcasting and positional encodings. Alternative architectures based on pre-trained Vision Transformers or latent diffusion models have also been proposed~\citep{Wu2023SlotDiffusionOG,Pramanik2024MaskedMS}.

For each image, the Slot Attention module receives a set of $N$ feature vectors in $\mathbb{R}^d$ and uses an iterative attention mechanism to refine a set of $K$ slots. To formalize this process, let $z \in \mathcal{Z}$ denote the input set of features, and let $s^{(t)} \in \mathcal{S}^K \subseteq \mathbb{R}^{K \times d}$ denote the slot representations at iteration $t$. Let $W_q, W_k, W_v \in \mathbb{R}^{d \times d}$ denote the query, key, and value projection matrices, respectively.

At iteration $t$, the attention weights $A^{(t)}$ and slot update $u^{(t)}$ are computed as
\[ u^{(t)} = B^{(t)} (z W_v), \quad \text{where} \quad B_{ij} \coloneqq \frac{A_{ij}}{\sum_k A_{ik}}, \quad A^{(t)} = \mathsf{Softmax} \left( \frac{(s^{(t)} W_q) (z W_k)^\top}{\sqrt{d}} \right) \in \mathbb{R}^{K \times N}. \]
The slots are updated according to $s^{(t+1)} = \mathsf{SlotUpdate}(s^{(t)}, u^{(t)})$, which consists of a Gated Recurrent Unit (GRU) followed by a multilayer perceptron (MLP) with a residual connection. The initial slots $s^{(0)}$ are randomly sampled from a learned Gaussian distribution. This iterative refinement is performed for $T$ rounds to produce the final slots $s^{(T)}$.

We next detail the additive slot decoder used to reconstruct the image from the set of slots. The decoder maps each slot to a masked image space $\mathcal{X}_m \subseteq \mathbb{R}^{H \times W \times (C+1)}$, where the image consists of $C$ channels together with an additional unnormalized alpha channel. Let $h_\theta : \mathcal{S} \to \mathcal{X}_m$ denote a per-slot decoder that maps each slot to an object in the image. 

Given slots $s = \{ s_1, \ldots, s_K \}$, let $\alpha_1, \ldots, \alpha_K$ denote the normalized alpha masks obtained by applying a softmax across slots. Then,
\[ f_\theta(s) = \sum_{i=1}^K \alpha_i h_\theta(s_i). \]
Hence, we refer to this architecture as an \emph{additive decoder}. The full training procedure for Slot Attention is outlined in Algorithm~\ref{alg:slot_attn_training}. For simplicity, we omit the batch dimension, but the algorithm can be easily modified to accommodate minibatches. We refer the reader to~\citet{Locatello2020ObjectCentricLearning} for additional details.

At this point, the algorithmic similarities between MPM and Slot Attention should begin to emerge. Like MPM, Slot Attention uses a bi-level scheme in which the slots are iteratively refined for each instance during training. We show that when viewed as clustering in the latent space, the model is effectively maximizing a bound on the data likelihood, thus providing a principled explanation for its effectiveness.

\section{TRAINING DETAILS} \label{app:training}
For object-centric image modeling, we adopt a convolutional neural network (CNN) architecture for both the generative model and the recognition network, following an encoder–decoder style design. Models are optimized with Adam using an initial learning rate of $4 \times 10^{-4}$ and step-based learning rate decay, which we find produces stable convergence across runs. We train for $1{,}000$ epochs, which requires approximately one hour for our model, and twice as long for Slot Attention.

The sequential text model is parameterized as a multinomial distribution over tokens, conditioned on a topic embedding, produced by a three-layer MLP. The recognition network uses a frozen pre-trained BERT model~\citep{Devlin2019BERTPO}, followed by a trainable two-layer MLP to generate contextual embeddings for each token. Word-level embeddings are obtained by averaging subword token embeddings, and articles are truncated to 512 tokens to align with BERT’s maximum input length. We use the Adam optimizer with an initial learning rate of $1 \times 10^{-5}$ and step-based learning rate decay, over $200$ epochs. The training requires roughly 1.5 hours.

\begin{figure}[t!]
    \centering
    \begin{subfigure}[t]{0.49\linewidth}
        \centering
        \includegraphics[width=\linewidth]{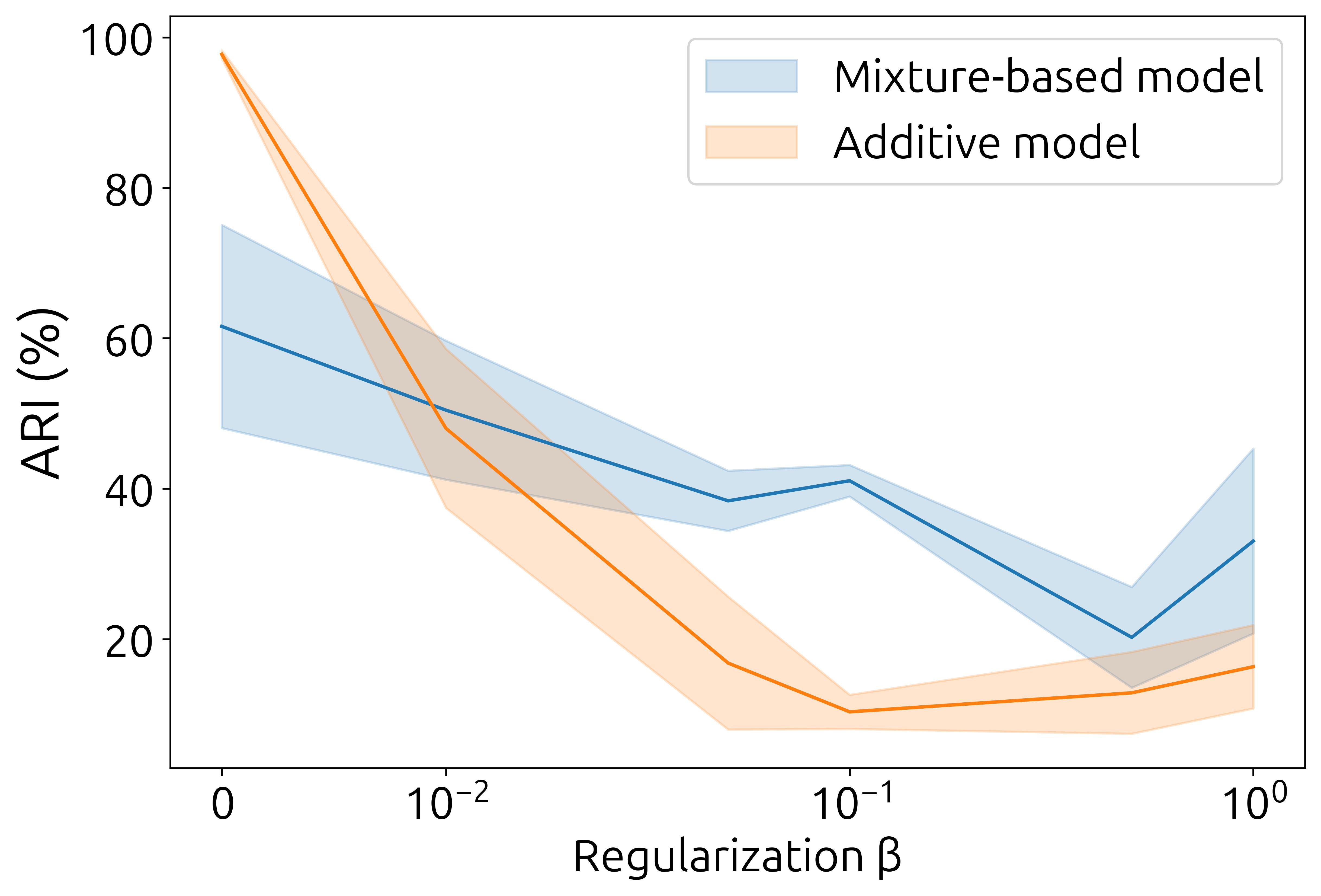}
        \caption{}
        \label{fig:beta_ocl}
    \end{subfigure}
    \hfill
    \begin{subfigure}[t]{0.49\linewidth}
        \centering
        \includegraphics[width=\linewidth]{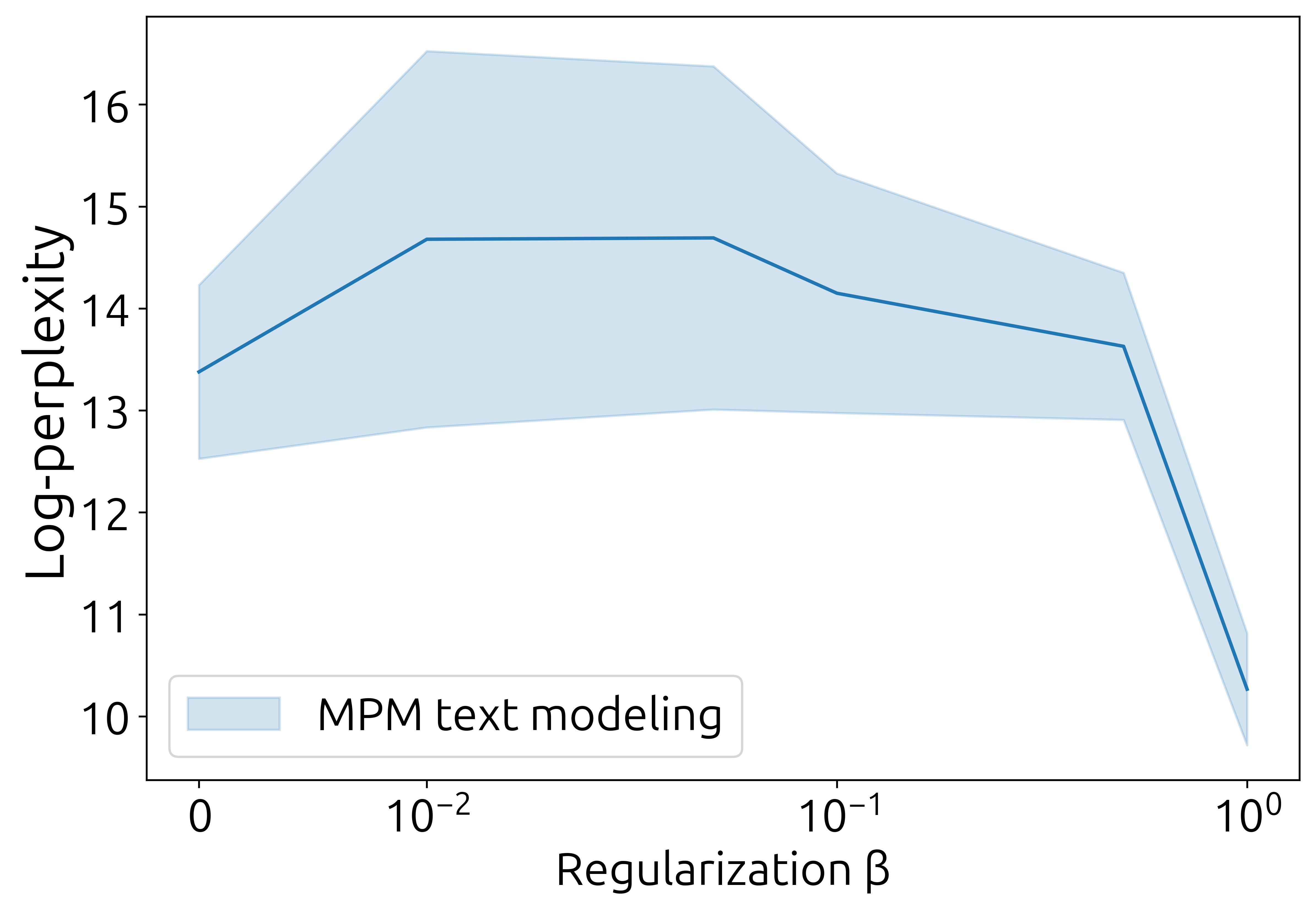}
        \caption{}
        \label{fig:beta_text}
    \end{subfigure}
    \caption{
    For the object-centric learning meta-probabilistic models, ARI decreases sharply as the regularization strength increases (Figure~\ref{fig:beta_ocl}). In contrast, the log-perplexity of our model in the text experiments remains mostly stable across values of $\beta$ (Figure~\ref{fig:beta_text}). For each value of $\beta$, we run five trials with different random initializations and training splits and report the mean and standard deviation.}
    \label{fig:beta}
\end{figure}

All experiments were performed on a single NVIDIA RTX 5070 GPU with 16GB memory. We tune learning rates via grid search over $\{1 \times 10^{-5}, 4 \times 10^{-5}, 1 \times 10^{-4}, 4 \times 10^{-4}, 1 \times 10^{-3}\}$. The hyperparameter $\beta$ is selected to be as large as possible from $\{0.01, 0.05, 0.1, 0.5, 1.0\}$ without significantly degrading reconstruction quality.

\section{ADDITIONAL EXPERIMENTS}

\subsection{Effect of the regularization parameter} \label{app:beta}

We examine how the regularization parameter $\beta$ influences model performance. Specifically, we vary $\beta$ logarithmically from $10^{-2}$ to $1$, and additionally evaluate the unregularized case $\beta = 0$. Figure~\ref{fig:beta} reports the ARI for the mixture and additive decoder models, along with the log-perplexity of the sequential text model. In the object-centric learning setting, performance degrades sharply as $\beta$ increases. This behavior arises because stronger regularization encourages the posterior distribution to become more uniform, which suppresses the underlying clustering structure. In contrast, for the sequential text modeling task, performance remains relatively stable across the range of $\beta$, with a modest improvement observed at $\beta = 1$.

\begin{figure}[t]
\centering
\begin{minipage}{0.484\textwidth}
    \centering
    \includegraphics[width=0.99\linewidth]{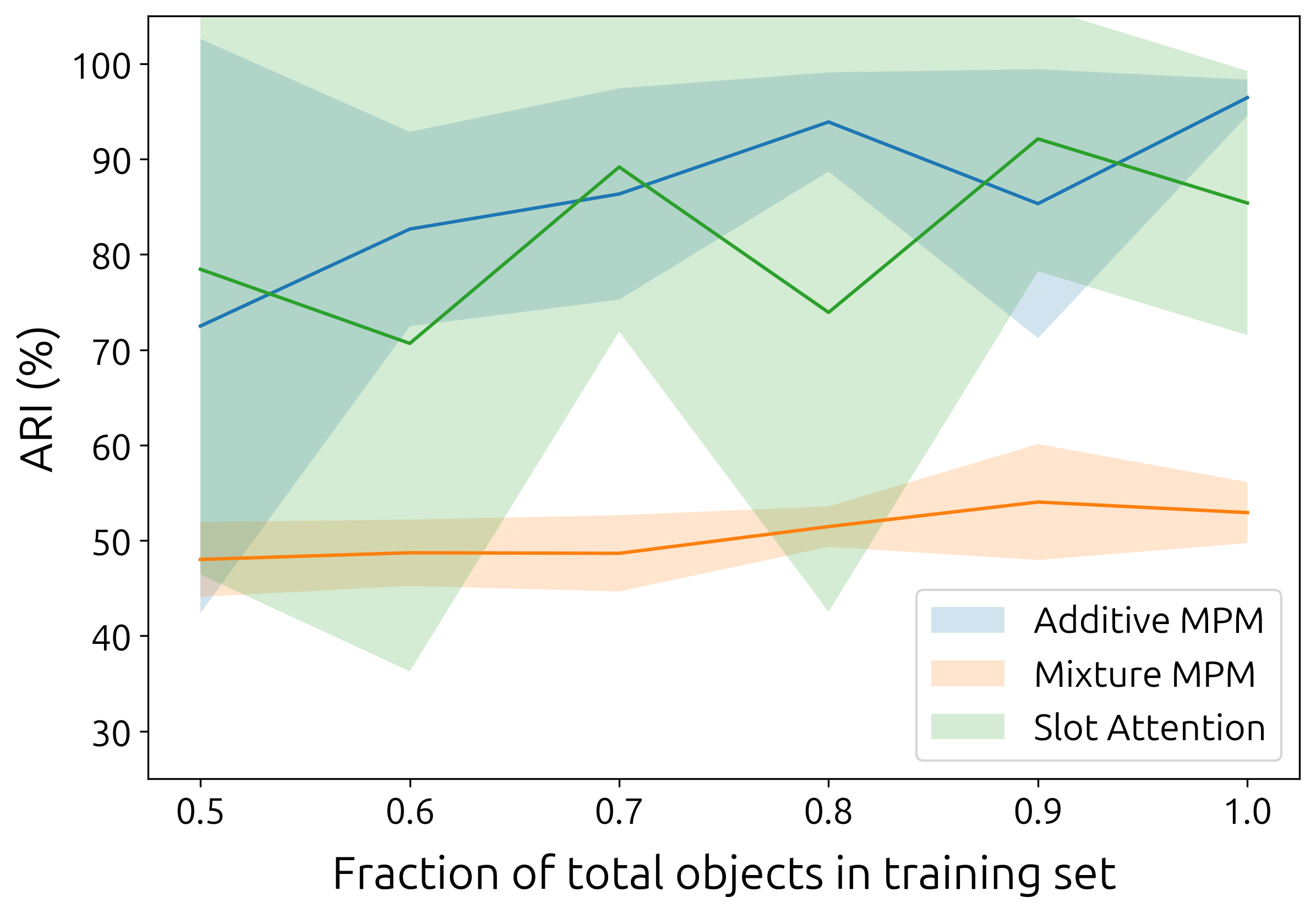}
    \caption{Object-centric learning performance gradually declines as task diversity decreases, as measured by the fraction of objects seen in the training data.}
    \label{fig:task-diversity}
\end{minipage}
\hfill
\begin{minipage}{0.461\textwidth}
    \centering
    \renewcommand{\arraystretch}{1.3}
    \setlength{\tabcolsep}{7.5pt}
    \captionof{table}{MPM achieves mean NPMI scores comparable to the baselines, while attaining competitive maximum NPMI values. We report the mean and standard deviation across five runs with different random seeds and initializations.}
    {\small
    \begin{tabular}{ccc}
        \toprule
        Model & NPMI (mean) & NPMI (max) \\
        \midrule
        LDA & $0.165 \pm 0.005$ & $0.599 \pm 0.145$\\
        NVDM & $0.070 \pm 0.002$ & $0.157 \pm 0.011$ \\
        GSM & $0.145 \pm 0.003$ & $0.424 \pm 0.041$  \\
        NTM & $0.217 \pm 0.011$ & $0.465 \pm 0.045$ \\
        NTMR & $0.072 \pm 0.001$ & $0.143 \pm 0.016$ \\
        WeTe & \underline{$\mathbf{0.230 \pm 0.003}$} & $0.617 \pm 0.028$ \\
        FASTopic & $0.207 \pm 0.004$ & \underline{$\mathbf{0.724 \pm 0.067}$} \\
        MPM (Ours) & $0.138 \pm 0.022$ & $0.624 \pm 0.181$ \\
        \bottomrule
    \end{tabular}}
    \label{tab:npmi}
\end{minipage}
\end{figure}

\subsection{Effect of task diversity on object-centric learning}
Meta-learning algorithms typically rely on the assumption that the underlying tasks exhibit sufficient diversity. As with other meta-learning approaches, our model also depends on the presence of meaningful cross-dataset patterns in order to learn transferable global structure.

To study how our model behaves under varying levels of task diversity, we use the number of distinct objects present in the training data as a proxy for task diversity. Each object corresponds to a tetromino characterized by a specific color, shape, and orientation. We construct restricted training datasets by sampling a random fraction of the full set of possible objects and retaining only images that contain objects from this subset. The fraction of allowed objects is varied over $\{0.5, 0.6, 0.7, 0.8, 0.9, 1.0\}$, where including all objects corresponds to the original experimental setting. Because restricting the object set reduces the number of available training images, we apply rotation and flip augmentations to keep the training set size consistent across settings.

For each level of task diversity, we evaluate across five random object restrictions and training seeds. Figure~\ref{fig:task-diversity} reports the mean and standard deviation of the ARI scores. As expected, all models exhibit a decline in ARI as the number of observed objects in the training data decreases, reflecting the reduced task diversity. The additive MPM and Slot Attention models show a gradual performance decline due to their compositional behavior, while the mixture model remains relatively stable across different levels of task diversity.

We note that restricting object types introduces a distribution shift between the training and test data, since the test set may contain objects that are absent from the training images. Nonetheless, the number of allowed object types serves as a useful proxy for task diversity: when fewer object types are available, the resulting training images become more similar to one another, effectively reducing the diversity of tasks.

\subsection{NPMI topic coherence results} \label{app:npmi}

In addition to evaluating our sequential text model using the UMass coherence score, we also report the Normalized Pointwise Mutual Information (NPMI) in Table~\ref{tab:npmi}. NPMI measures the co-occurrence frequency between top words within a topic and normalizes using log-probability. Our meta-probabilistic model achieves a maximum NPMI coherence comparable to existing neural topic modeling approaches, including WeTe and FasTOPIC. However, the mean NPMI score is lower, indicating that while some topics exhibit strong coherence, the overall performance remains within the mid-range of the evaluated models.

\subsection{Scaling behavior of MPM text model} \label{app:scaling}

We conduct a study based on the text experiment to directly examine how the runtime and memory of our meta-probabilistic model scale with data dimensionality. Specifically, we vary the maximum input sequence length and report the resulting computational cost in Table~\ref{tab:scaling} for our meta-probabilistic model (MPM) and for fine-tuning BERT (BERT-FT) as a baseline. Training times for MPM roughly align with BERT-FT, while there is an increase in GPU memory usage due to the additional computation graph introduced by the inner optimization. 

\begin{table}[t]
    \centering
    \caption{When using BERT as the contextual word embedding model in the text experiments, our meta-probabilistic models incur a similar runtime cost to fine-tuning BERT, as the inner optimization is lightweight. However, memory usage is higher than standard fine-tuning due to the additional optimization steps.}
    \begin{tabular}{ccccc}
    \toprule
    Max sequence length & Time (MPM) & Time (BERT-FT) & GPU (MPM) & GPU (BERT-FT) \\
    \midrule
    64  & 0.3 hr & 0.2 hr & 3 GB  & 2 GB \\
    128 & 0.4 hr & 0.3 hr & 5 GB  & 3 GB \\
    256 & 0.6 hr & 0.5 hr & 10 GB & 4 GB \\
    512 & 1.2 hr & 1.0 hr & 16 GB & 7 GB \\
    \bottomrule
    \end{tabular}
    \label{tab:scaling}
\end{table}

In general, we find that the scaling behavior of our models largely depends on the choice of the underlying models $f_\theta$ and $g_\phi$. Empirically, we find that the runtime is dominated by these components, since the bilevel optimization overhead is relatively small. On the other hand, memory usage can exhibit more noticeable increases because intermediate activations in the iterative procedure are retained.

\section{EXPERIMENTAL ASSETS} \label{app:assets}
All code for our meta-probabilistic models are publicly available.\footnote{\url{https://github.com/kzhangm02/mpm}} In the experiments, we use the publicly available Tetrominoes~\citep{multiobjectdatasets19} dataset, distributed under the Apache License, and AP News corpus~\citep{Harman1993OverviewTREC} dataset. For the text experiments, we follow the original LDA work~\citep{Blei2003LatentDA} by using a subset of the AP News corpus, which is available under the GNU Lesser General Public License.\footnote{\url{https://github.com/blei-lab/lda-c}} For comparison with Slot Attention, we use our own implementation based on a publicly available version.\footnote{\url{https://github.com/evelinehong/slot-attention-pytorch}}

\end{document}